\documentclass[11pt,a4paper]{article}
\usepackage[margin=20mm,headheight=14pt]{geometry}
\usepackage[T1]{fontenc}
\usepackage[utf8]{inputenc}
\usepackage{lmodern}
\usepackage{microtype}
\usepackage{amsmath,amssymb}
\usepackage{graphicx}
\usepackage{subcaption}
\usepackage{booktabs,array,tabularx,longtable}
\usepackage{xcolor}
\usepackage{placeins}
\usepackage{enumitem}
\usepackage[numbers,sort&compress]{natbib}
\usepackage{url}
\usepackage{hyperref}
\usepackage{fancyhdr}
\usepackage{titlesec}
\hypersetup{hidelinks,pdftitle={Beyond Benchmark Scores: Auditing Medical Vision-Language Models for Chest X-Ray Tuberculosis Screening},pdfauthor={}}
\definecolor{academicblue}{HTML}{1F4E79}
\definecolor{paletteolive}{HTML}{7D8F69}
\definecolor{palettesage}{HTML}{A7C1A8}
\definecolor{palettecream}{HTML}{E8D5C4}
\definecolor{palettecoral}{HTML}{C46A5D}
\newcommand{\headercell}[1]{\textcolor{academicblue}{\textbf{#1}}}
\newcommand{\question}[1]{\textcolor{academicblue}{\textbf{#1}}}
\titleformat{\section}{\large\bfseries\color{academicblue}}{\thesection}{0.65em}{}
\titleformat{\subsection}{\normalsize\bfseries\color{academicblue}}{\thesubsection}{0.65em}{}
\titleformat{\subsubsection}{\normalsize\bfseries}{\thesubsubsection}{0.65em}{}
\title{\LARGE\bfseries Beyond Benchmark Scores: Auditing Medical Vision-Language Models for Chest X-Ray Tuberculosis Screening}
\author{
\mbox{Mushir Akhtar \quad M.~Tanveer \quad Mohd.~Arshad}\\[4pt]
\small Department of Mathematics, Indian Institute of Technology Indore, India\\[3pt]
\small
\href{mailto:mushirakhtar.ml@gmail.com}{\texttt{mushirakhtar.ml@gmail.com}}
\quad
\href{mailto:mtanveer@iiti.ac.in}{\texttt{mtanveer@iiti.ac.in}}
\quad
\href{mailto:arshad@iiti.ac.in}{\texttt{arshad@iiti.ac.in}}
}
\date{}

\begin{document}
\maketitle
\vspace{-1.8em}
\begin{abstract}
\noindent
A medical model's benchmark score does not establish that the same conclusion holds under a different evaluation. This study tests whether claims about model ranking, score reliability and screening performance survive changes in cohort, prompt, negative spectrum, specified prevalence and operating threshold. We audit three medical vision-language models (BioMedCLIP, CheXficient, and MedSigLIP) and a general-domain OpenCLIP comparator on 12,200 chest radiograph records from four datasets (Montgomery, Shenzhen, TBX11K, and VinDr-CXR). Five fixed prompt families yield 244,000 model--image--prompt scores. No model leads every cohort and reliability criterion. Prompt-family changes alter AUROC in 21 of 48 multiplicity-controlled comparisons. Replacing healthy controls with sick non-tuberculosis controls reduces AUROC by 0.075--0.306 across all four models. On VinDr-CXR, the three medical models distinguish tuberculosis from no-finding controls substantially better than from pneumonia or lung tumor; their AUROC point estimates for both named diseases fall below 0.5. CheXficient has documented VinDr-CXR pretraining exposure, which limits the interpretation of its results. Thresholds chosen for 95\% sensitivity on TBX11K training retain that constraint by point estimate in only four of sixteen target evaluations. A five-seed supervised source model reaches 0.999 AUROC on TBX11K validation but 0.629 on each of two external cohorts. Conservative exclusion of perceptual-overlap candidates narrows this gap without closing it. These retrospective, single-task results show that discrimination, score reliability and threshold retention support different portability claims. Evidence for chest X-ray tuberculosis screening should identify the complete evaluation specification rather than attribute clinical portability to a checkpoint alone.
\end{abstract}

\section{Introduction}

A model can rank tuberculosis cases accurately in one chest X-ray benchmark and still provide weak evidence for screening elsewhere. A high AUROC does not specify which non-tuberculosis patients serve as controls, how text defines the classifier, or whether an unchanged threshold identifies enough cases in another population. These distinctions matter when benchmark results inform decisions about clinical reuse.

Medical vision-language models make reuse possible without fitting a new classifier to local labels. Image and text encoders map a radiograph and a diagnostic description into a shared representation. Their similarity supplies a task-specific score. This approach extends the transfer capabilities of general vision-language learning to biomedical images and chest radiography \citep{radford2021learning,tiu2022chexzero,zhang2023biomedclip}. Its flexibility also changes the object under evaluation. A checkpoint does not determine the classifier by itself: the prompt, image processor and scoring rule participate directly in the prediction.

The consequences of distribution shift are well established. Chest radiograph models can exploit site-specific information and generalize unevenly across hospitals \citep{zech2018variable}. External validation cannot guarantee validity at future sites, which motivates recurring local assessment \citep{youssef2023external}. The question here is not simply whether performance changes under domain shift, but \emph{which reported conclusion survives a declared change in the evaluation}. A model may retain strong discrimination while losing a sensitivity constraint. It may look reliable against normal lungs but rank another disease above tuberculosis. Two models with similar AUROC may also differ in score reliability or selective prediction.

Existing work addresses several parts of this problem. Prompt studies examine how text alters zero-shot medical classification \citep{wang2024prompts}. FairMedFM evaluates subgroup fairness, MediConfusion probes difficult clinical alternatives, and medical-image artifact studies test robustness to image degradation \citep{jin2024fairmedfm,sepehri2025mediconfusion,cheng2025artifacts}. Hidden stratification explains why aggregate scores can conceal clinically meaningful failures \citep{oakdenrayner2020hidden}. Recent perspectives emphasize specification-dependent robustness and evaluation grounded in clinical context \citep{xian2025robustness,bielick2026benchmarks}. Together, this work motivates an evaluation in which each claim is linked to the conditions that support it.

We represent the evaluation specification as
\begin{equation}
  \mathcal{E}=(m,p,C,N,\pi,\tau,r),
  \label{eq:specification}
\end{equation}
where $m$ denotes the checkpoint and image processor, $p$ the text and aggregation rule, $C$ the sampled cohort, $N$ the negative-class definition, $\pi$ a specified prevalence, $\tau$ the decision threshold, and $r$ the performance criterion. This notation records the conditions attached to a claim; it is not a new predictive model. Portability means retention of the stated conclusion under a specified change, rather than numerical invariance of every metric.

Four questions guide the study. \question{Q1: Does the leading model remain the same across cohorts and performance criteria?} \question{Q2: How much do prompt specification and the negative spectrum change discrimination when the images or positive cases remain fixed?} \question{Q3: Do score reliability and source-derived operating thresholds remain adequate under new prevalence assumptions and target cohorts?} \question{Q4: Does strong supervised source validation resolve the transfer problem, and how sensitive is that conclusion to potential cross-cohort image overlap?}

We address these questions using four vision-language models, four chest X-ray datasets and five prompt families. Paired comparisons separate prompt and negative-spectrum effects from changes in the positive sample. Prevalence standardization examines score behavior under explicit class weights. Threshold transport preserves a source decision rule at the target. Five-seed supervised training and conservative overlap-candidate exclusion provide complementary evidence about source validation. The resulting audit links model-ranking, reliability and screening claims to specified changes in evaluation, while preserving the uncertainty and exposure qualifications attached to each comparison.

The remainder of the paper is organized as follows. Section~\ref{sec:methods} defines the audit design, evaluation specifications, and statistical analyses. Section~\ref{sec:results} reports the cohort, prompt, negative-spectrum, prevalence, threshold-transport, and supervised-transfer results. Section~\ref{sec:discussion} interprets their implications and limitations, and Section~\ref{sec:conclusion} summarizes the evidence for evaluating portability claims.

\section{Audit Design and Methods}
\label{sec:methods}

\subsection{Study scope and analysis sequence}

The study is a retrospective benchmark audit of image-level tuberculosis (TB) labels. It includes 12,200 image records: 9,200 from Montgomery, Shenzhen and the labeled TBX11K training and validation splits, followed by a 3,000-image VinDr-CXR test extension. Four models and five prompt families produce 244,000 scores. The unit is an image record, not a verified unique patient across datasets. Potential cross-cohort similarity is examined separately in Section~\ref{sec:overlap-methods}.

The primary zero-shot comparison uses the clinical prompt family on Montgomery, Shenzhen and TBX11K validation. TBX11K training supplies a secondary zero-shot replication cohort and the primary source for threshold estimation. The supervised experiment partitions that training split into optimization and checkpoint-selection subsets. VinDr-CXR extends the audit to a larger, radiologist-annotated cohort with named alternative diagnoses.

We conduct the analyses in sequential blocks. Each block fixes its cohort rules, prompts, scoring implementation and analysis configuration before outcome tables are assembled. This design prevents target-specific prompt or threshold tuning within the reported comparisons, but does not constitute prospective study-wide preregistration. The all-prompt score average, variance decomposition and overlap-candidate sensitivity are secondary analyses. No target cohort selects the primary prompt or a supervised checkpoint. The five supervised models are trained on the source data; the zero-shot vision-language checkpoints remain frozen throughout.

\subsection{Cohorts and reference labels}
\label{sec:cohorts}

Table~\ref{tab:cohorts} defines the evaluated cohorts. Montgomery and Shenzhen provide TB-positive and normal chest radiographs \citep{jaeger2014tb}. TBX11K distinguishes TB, healthy and sick non-TB categories \citep{liu2020tbx11k}. We use its official labeled training and validation lists. The unlabeled official test split and the additional datasets bundled with the download are excluded. In particular, bundled copies of the NLM datasets do not enter the TBX11K manifest.

\begin{table}[!htbp]
\centering
\caption{Cohort composition and experimental role.}
\label{tab:cohorts}
\small
\setlength{\tabcolsep}{4pt}
\begin{tabular}{llrrrl}
\toprule
\headercell{Cohort} & \headercell{Non-TB composition} & \headercell{TB} & \headercell{Non-TB} & \headercell{TB (\%)} & \headercell{Role} \\
\midrule
Montgomery & Normal & 58 & 80 & 42.0 & External comparison \\
\midrule
Shenzhen & Normal & 336 & 326 & 50.8 & External comparison \\
\midrule
TBX11K validation & 800 healthy; 800 sick & 200 & 1,600 & 11.1 & Primary benchmark \\
\midrule
TBX11K training & 3,000 healthy; 3,000 sick & 600 & 6,000 & 9.1 & Source and replication \\
\midrule
VinDr-CXR test & No finding; other findings & 164 & 2,836 & 5.5 & Named-spectrum extension \\
\bottomrule
\end{tabular}
\par\smallskip
\begin{minipage}{.98\linewidth}\footnotesize
Counts describe image records rather than verified unique patients across cohorts. ``Sick'' denotes the dataset's sick non-TB category. VinDr-CXR diagnoses can co-occur; its all-non-TB group includes the full set of TB-negative records, not only the named diagnoses used in secondary contrasts.
\end{minipage}
\end{table}

VinDr-CXR version 1.0.0 contributes the 3,000-image test set from two Vietnamese hospitals \citep{nguyen2022vindr,nguyen2021vindrphysionet,goldberger2000physionet}. We use the global image-level annotations, not a new reader study. The dataset describes test annotations obtained by consensus among five radiologists. The downloaded labels identify 164 TB-positive records, 2,051 with no finding, 246 with pneumonia, 80 with lung tumor and two with COPD. These totals are not mutually exclusive. To construct negative groups, we retain a named diagnosis only when the image is TB-negative. This yields 210 pneumonia and 73 lung-tumor controls. COPD remains descriptive because its sample is too small for meaningful comparative inference.

The endpoint follows each dataset's TB label. The datasets do not provide one harmonized microbiological reference standard for this audit. In particular, VinDr-CXR labels describe radiographic disease impressions. We therefore interpret the results as agreement with benchmark TB labels in a screening-oriented task, not as prospective diagnostic accuracy for bacteriologically confirmed TB. Differences in acquisition, labeling and disease severity remain part of the cohort comparison.

\subsection{Image processing and checkpoint identity}
\label{sec:implementation}

All original raster images are converted to RGB before the checkpoint's image processor. VinDr-CXR DICOM files require an additional, fixed conversion. The conversion reads the pixel array, applies the modality transformation and an available VOI transformation, and excludes padding and non-finite pixels from intensity-range estimation. Valid intensities undergo per-image min--max scaling. MONOCHROME1 images are inverted; MONOCHROME2 images retain their polarity. Invalid pixels map to zero. The resulting image is rounded to 8-bit grayscale and stored as PNG before RGB conversion. All 3,000 test images pass preparation. No lung segmentation, target-specific crop or target-label-dependent intensity adjustment is applied.

The four models cover distinct training regimes (Table~\ref{tab:models}). OpenCLIP provides a general-domain comparator rather than a fourth medical model. BioMedCLIP learns from biomedical figure--caption pairs. CheXficient specializes in chest radiography. MedSigLIP uses medical image--text training across several specialties. Each zero-shot experiment retains the checkpoint weights and learned similarity scale. The named checkpoint processor determines image resizing and normalization.

\begin{table}[!htbp]
\centering
\caption{Vision-language checkpoints and the interpretation of their training provenance.}
\label{tab:models}
\small
\setlength{\tabcolsep}{4pt}
\begin{tabular}{>{\raggedright\arraybackslash}p{.13\linewidth}>{\raggedright\arraybackslash}p{.38\linewidth}>{\raggedright\arraybackslash}p{.41\linewidth}}
\toprule
\headercell{Model} & \headercell{Checkpoint and representation} & \headercell{Pretraining context and exposure qualification} \\
\midrule
OpenCLIP & ViT-B/32; \texttt{laion2b\_s34b\_b79k}; $224\times224$ input & General image--text training on LAION-2B. No documented direct dataset exposure in the examined documentation; web-image overlap remains unresolved \citep{cherti2023openclip}. \\
\midrule
BioMedCLIP & \texttt{microsoft/BiomedCLIP-}\newline\texttt{PubMedBERT\_256-vit\_base\_}\newline\texttt{patch16\_224}; $224\times224$ input & Biomedical figure--caption training on PMC-15M. Scientific-figure reuse prevents certification of image-level independence \citep{zhang2023biomedclip}. \\
\midrule
CheXficient & \texttt{StanfordAIMI/CheXficient}; DINOv2-base and BioClinicalBERT; checkpoint image processor & CXR-specific training. VinDr-CXR is a documented training source; TBX11K is described as unseen in the model report. The VinDr experiment is a within-dataset stress test, not independent external validation \citep{wang2026chexficient}. \\
\midrule
MedSigLIP & \texttt{google/medsiglip-448}; medical SigLIP image and text encoders; $448\times448$ input & Medical training across specialties, including public and non-public sources. No documented direct exposure to the evaluated datasets; absence of exposure is not established \citep{sellergren2025medgemma,google2025medsiglipcard}. \\
\bottomrule
\end{tabular}
\end{table}

CheXficient uses repository revision \texttt{34bb5ea905c4cb504d2b73c9a6687a5d5082f52a}; MedSigLIP uses \texttt{9cea28a1a1195f665105faa6e8544c112fd960a4}. The BioMedCLIP identifier and OpenCLIP weight tag specify their selected weights, but the original runs do not provide the same explicit immutable-revision record for every checkpoint. This is a provenance limitation. Execution uses NVIDIA A100 40-GB GPUs with PyTorch 2.5.1 and CUDA 12.4. OpenCLIP inference uses OpenCLIP 3.3.0; the CheXficient and MedSigLIP runners use Transformers 4.51.3. VinDr conversion uses pydicom 3.0.2.

MedSigLIP preprocessing uses the Hugging Face AutoProcessor with its PIL-based resize, not the TensorFlow/Big Vision resize used for reference results in its model card \citep{google2025medsiglipcard}. The present estimates consequently characterize this specified implementation and are not exact reproductions of the model developer's benchmark numbers.

\subsection{Prompt-defined classifiers}
\label{sec:scoring}

Table~\ref{tab:prompts} gives every prompt verbatim. The primary clinical family contains four TB descriptions and four non-TB descriptions. The four alternatives use one positive--negative pair each. Their differences include syntax, the explicit negative concept and ensemble size. In particular, a normal-radiograph prompt is not semantically identical to absence of TB in an otherwise abnormal image. We therefore call this a \emph{prompt-family sensitivity analysis}, rather than attributing every effect to a small paraphrase.

For checkpoint $m$, let $v_m(x)$ denote the unit-normalized image embedding. Let $u_m(t)$ denote a unit-normalized text embedding for text $t$. The prototype for class $c\in\{0,1\}$ under prompt family $p$ is
\begin{equation}
  z_{m,p,c}=\frac{1}{|T_{p,c}|}\sum_{t\in T_{p,c}}u_m(t),
  \qquad
  q_{m,p,c}=\frac{z_{m,p,c}}{\|z_{m,p,c}\|_2}.
\end{equation}
Using the checkpoint's learned positive scale $\alpha_m$, the TB score is
\begin{equation}
 s_{m,p}(x)=
 \frac{\exp\{\alpha_m v_m(x)^\top q_{m,p,1}\}}
 {\exp\{\alpha_m v_m(x)^\top q_{m,p,0}\}+\exp\{\alpha_m v_m(x)^\top q_{m,p,1}\}}.
 \label{eq:score}
\end{equation}
This two-class score compares TB text with its specified negative text. It is not a calibrated probability of disease. MedSigLIP's shared logit bias cancels in this normalization. Its score here is therefore a relative two-class softmax, not its native independent sigmoid probability for one text. The learned scale remains in the score, so ranking and score magnitude can respond differently to the prompt.

Within a model--cohort run, all five families use the same encoded images. No target label selects the primary family or the highest-scoring alternative. A secondary analysis averages the five family-level probabilities with equal weights. This differs from averaging text embeddings within the clinical family. We compare the score average with the primary classifier without choosing weights from target outcomes.

\begin{table}[!htbp]
\centering
\caption{Exact text supplied to each frozen text encoder.}
\label{tab:prompts}
\small
\setlength{\tabcolsep}{4pt}
\renewcommand{\arraystretch}{1.12}
\begin{tabular}{>{\raggedright\arraybackslash}p{.10\linewidth}>{\raggedright\arraybackslash}p{.41\linewidth}>{\raggedright\arraybackslash}p{.41\linewidth}}
\toprule
\headercell{Family} & \headercell{TB text} & \headercell{Non-TB text} \\
\midrule
Clinical & a chest x-ray showing pulmonary tuberculosis\newline a chest radiograph with tuberculosis\newline a chest x-ray with findings compatible with tuberculosis\newline a frontal chest radiograph showing active tuberculosis & a normal chest x-ray without tuberculosis\newline a chest radiograph with no evidence of tuberculosis\newline a chest x-ray without pulmonary tuberculosis\newline a frontal chest radiograph without signs of tuberculosis \\
\midrule
Photo & this is a photo of a chest X-ray showing pulmonary tuberculosis & this is a photo of a chest X-ray without pulmonary tuberculosis \\
\midrule
Report & chest radiograph findings are compatible with pulmonary tuberculosis & no radiographic evidence of pulmonary tuberculosis \\
\midrule
Presence & pulmonary tuberculosis is present on this chest radiograph & pulmonary tuberculosis is absent on this chest radiograph \\
\midrule
Normal & a chest radiograph showing pulmonary tuberculosis & a normal chest radiograph \\
\bottomrule
\end{tabular}
\par\smallskip
\begin{minipage}{0.98\linewidth}\footnotesize Clinical averages four normalized text embeddings per class, then renormalizes the mean. Other families use one text per class. Capitalization matches the executed configuration.\end{minipage}
\end{table}

\FloatBarrier

\subsection{Paired negative-spectrum comparisons}
\label{sec:negative-methods}

TBX11K validation permits a controlled change of negative class. We score the same 200 TB images against either 800 healthy or 800 sick non-TB images. The primary estimand for model $m$ is
\begin{equation}
 D_m=\operatorname{AUROC}_m(P,H)-\operatorname{AUROC}_m(P,S),
 \label{eq:negative}
\end{equation}
where $P$ contains the shared positive cases, $H$ the healthy controls and $S$ the sick controls. Both contrasts have 20\% TB prevalence. The comparison therefore changes the negative spectrum without changing the positive sample or class balance. It does not identify which disease categories account for the sick-control effect.

VinDr-CXR supplies the named-disease extension. We compare the same 164 TB-positive images with no-finding, pneumonia and lung-tumor controls. The estimand is named-negative AUROC minus no-finding AUROC. Negative group sizes differ, and named diagnoses can overlap. The contrasts consequently do not describe mutually exclusive disease populations. They also do not remove other findings from the positive TB group.

\subsection{Discrimination, score reliability and selective prediction}
\label{sec:metrics}

AUROC estimates the probability that a positive image receives a higher score than a negative image, with half credit for ties. AUPRC is average precision, computed by summing precision at positive rank increments without trapezoidal interpolation. Equal scores are ordered at individual ranks; pooling tied thresholds changes the clinical-prompt point estimates by less than $10^{-5}$ across all twenty model--cohort cells, including TBX11K training. AUROC uses average ranks for ties. AUPRC remains secondary because its dependence on prevalence precludes interpreting every between-cohort change as altered discrimination.

The Brier score is $n^{-1}\sum_i(s_i-y_i)^2$. It measures overall probabilistic prediction error, which combines discrimination and calibration. It is not a pure calibration measure. Expected calibration error uses 15 equal-width score bins:
\begin{equation}
 \operatorname{ECE}_{15}=\sum_{b=1}^{15}\frac{|B_b|}{n}
 \left|\frac{1}{|B_b|}\sum_{i\in B_b}s_i-
 \frac{1}{|B_b|}\sum_{i\in B_b}y_i\right|.
\end{equation}
Empty bins contribute zero and the final bin includes score one. Calibration plots use ten equal-width bins for readability; their binning differs from ECE$_{15}$. Small bins can be unstable, especially in Montgomery. We retain both score-error and calibration summaries because discrimination alone does not determine the meaning of a numerical score \citep{guo2017calibration,vancalster2019calibration}.

For within-cohort operating points, we report the largest sensitivity among observed thresholds with specificity at least 0.95, and the largest specificity with sensitivity at least 0.95. These use the target ROC curve. They are distinct from applying a threshold estimated in another cohort.

Selective prediction orders images by confidence $\max(s_i,1-s_i)$, while the binary decision remains $\mathbf{1}\{s_i\geq0.5\}$. Risk at a given coverage is the fraction of incorrect decisions among the most confident retained images. We report risk at 80\% and 90\% coverage and the trapezoidal area under the empirical risk--coverage curve (AURC), from coverage $1/n$ to one \citep{geifman2017selective}. Lower risk and AURC are preferable. This is a fixed-score abstention analysis; it does not simulate a clinician's decisions on deferred cases or optimize a deployment threshold.

\subsection{Prevalence-standardized score assessment}
\label{sec:prevalence}

To distinguish the effect of class proportions from the observed score distributions within each class, positives receive total weight $\pi$ and negatives total weight $1-\pi$. For positive set $P$ and negative set $N$, the standardized Brier score is
\begin{equation}
 \operatorname{BS}_{\pi}=
 \frac{\pi}{|P|}\sum_{i\in P}(1-s_i)^2+
 \frac{1-\pi}{|N|}\sum_{i\in N}s_i^2.
 \label{eq:brier}
\end{equation}
Weighted ECE and log loss use the same weights. This calculation retains the sampled conditional distributions of scores. It changes neither the labels nor the scores and does not remove changes in disease severity or acquisition. It describes a specified reweighting, not a verified future deployment population.

The original cohorts use $\pi\in\{0.10,0.20,0.50\}$, with 0.50 designated as the primary standardized comparison in that block. VinDr-CXR uses $\pi\in\{0.05,0.10,0.20\}$, with 0.10 primary. These grids support within-block comparisons; different designated primary prevalences must not be conflated. No score recalibration is applied.

\subsection{Source-threshold transport}
\label{sec:threshold-methods}

The primary decision rule predicts TB when $s\geq\tau_C$, where $\tau_C$ is the highest observed source threshold achieving sensitivity of at least 0.95 in source cohort $C$. TBX11K training supplies the primary source. The selected threshold is carried unchanged to TBX11K validation, Shenzhen, Montgomery and VinDr-CXR. Four models across four targets yield sixteen transports. Shenzhen provides a secondary source for transfer to TBX11K validation and Montgomery. A secondary analysis maximizes source sensitivity subject to specificity of at least 0.95. Ties favor higher specificity, then the highest threshold.

For target $C'$, $\operatorname{Se}_{C'}(\tau_C)$ denotes sensitivity and $\operatorname{Sp}_{C'}(\tau_C)$ specificity at the source threshold. Sensitivity retention is
\begin{equation}
 R_{C\rightarrow C'}=\mathbf{1}\{\operatorname{Se}_{C'}(\tau_C)\geq0.95\}.
\end{equation}
The indicator equals one when the target point estimate reaches 0.95 and zero otherwise. It does not require the lower confidence bound to reach 0.95. The target specificity is essential to interpretation: predicting almost every image as TB can retain sensitivity without yielding a useful screening rule. The 95\% constraints are experimental operating points, not claims that this study satisfies a particular clinical screening standard.

\subsection{Supervised source model}
\label{sec:supervised-methods}

An ImageNet-1K V2 initialized ResNet-50 provides a conventional supervised reference \citep{he2016deep}. Its final layer has one binary output. A fixed split of TBX11K training allocates 80\% of each original class to optimization and 20\% to internal selection, using split seed 2026. Optimization contains 5,280 images: 480 TB, 2,400 healthy and 2,400 sick non-TB. Selection contains 1,320 images: 120 TB and 600 in each negative category. Official TBX11K validation remains separate from both subsets.

Five optimization seeds, 2026--2030, share this split. Training uses positive-class-weighted binary cross-entropy, with the negative-to-positive count ratio as the positive weight. AdamW uses learning rate and weight decay $10^{-4}$, batch size 64 and BF16 mixed precision. Training stops after at most 20 epochs or five epochs without improved selection AUROC. The checkpoint with the highest selection AUROC supplies predictions on every evaluation cohort.

Images have $224\times224$ model input. Training augmentation uses random resized crops with area fraction 0.85--1.00 and aspect ratio 0.95--1.05, rotations within five degrees, translations within 2\%, and brightness and contrast jitter of 0.10. Horizontal reflection is absent. Evaluation resizes the shorter side to 256 pixels and takes a 224-pixel center crop. Normalization uses the ImageNet channel means and standard deviations. The five-seed ensemble averages predicted probabilities, without fitting ensemble weights. This experiment tests one fixed source-training protocol; it does not establish the best attainable supervised TB classifier. It does not include supervised VinDr-CXR evaluation.

\subsection{Overlap-candidate sensitivity}
\label{sec:overlap-methods}

Cross-dataset comparisons in the original 9,200-image matrix use byte-level SHA256 hashes and a 64-bit perceptual hash. The latter converts images to grayscale, resizes to $32\times32$, and thresholds low-frequency discrete-cosine coefficients relative to their median. A Hamming distance of at most four identifies candidate pairs across different datasets. A limited accessible reference-figure search does not establish coverage of any complete pretraining corpus. VinDr-CXR is outside this image-overlap screen.

Perceptual similarity is a screening signal, not proof of duplication. No radiologist or independent image-identity adjudication resolves the candidate pairs. The sensitivity analysis therefore removes every implicated image, without classifying any pair as a confirmed duplicate or nonduplicate. It preserves the original optimization and selection roles among retained TBX11K training images, then repeats all five seeds under the same training configuration.

We distinguish two comparisons. Applying the original ensemble to full versus retained targets measures the effect of changing target composition. Applying the original and retrained ensembles to the same retained targets measures the retraining contrast. The latter avoids mixing a model change with a simultaneous target-sample change. It still does not estimate the causal effect of verified leakage removal: the exclusions are unadjudicated and alter the source sample as well as its potential overlap risk.

\subsection{Statistical inference and uncertainty scope}
\label{sec:statistics}

All bootstrap intervals use 2,000 replicates, percentile endpoints at 2.5\% and 97.5\%, and analysis seed 2026. The resampling design differs by analysis block. In the original model--cohort--prompt matrix, identifier-based resampling reduces to an \emph{ordinary image bootstrap}: the manifests contain one image per identifier. Class proportions can vary between replicates. Model and prompt differences use paired indices. The marginal TBX11K subgroup intervals use this same original design.

The later prevalence, threshold, direct negative-spectrum, candidate-exclusion and VinDr blocks use resampling \emph{within label}, preserving positive and negative sample sizes. For the direct TBX11K negative-spectrum contrast, the positive resample is shared across the healthy and sick comparisons, while negative groups are resampled separately. Two-sided tests permute negative-group membership 10,000 times. Holm correction covers the four primary model tests \citep{holm1979sequential}. VinDr named-spectrum differences similarly share the positive resample; their intervals are descriptive and do not belong to the TBX11K correction family.

The 48 primary prompt contrasts compare four alternatives against the clinical family across four models and three original primary cohorts. Paired DeLong tests \citep{delong1988roc} receive Benjamini--Hochberg false-discovery correction at 0.05 across all 48 tests \citep{benjamini1995fdr}. The count of corrected differences does not include VinDr-CXR or TBX11K training. Other model, prompt-average and score-reliability intervals remain descriptive unless a correction is explicitly stated. An interval containing zero does not establish equivalence.

For original threshold transports, each replicate resamples the source within label, re-estimates its threshold and applies that threshold to an independently resampled target. VinDr threshold intervals instead condition on the already fixed source threshold and resample only the target. These two intervals answer different uncertainty questions. The pooled sixteen-transport retention count uses point estimates, for which this distinction does not change the definition.

For supervised training, the standard deviation across five seeds describes optimization variability on a shared data split. It is not a sampling confidence interval. Candidate-exclusion paired intervals condition on the original and retrained five-seed ensembles and resample the retained target images within label. Seed-paired retraining effects are reported separately. The available manifests lack longitudinal patient links, so all image-level intervals may underrepresent uncertainty if records remain correlated within patients or acquisition groups.

\subsection{Exposure and claim boundaries}

Dataset provenance qualifies the meaning of external evaluation. CheXficient explicitly includes VinDr-CXR among its pretraining sources \citep{wang2026chexficient}. We mark its VinDr results as documented dataset exposure without asserting that the specific test images occur in its training set. The other model descriptions do not document direct use of these evaluation datasets, but large web, scientific-figure and non-public corpora prevent a complete membership audit. Absence of documentation is not evidence of independence.

Montgomery, Shenzhen, TBX11K and VinDr-CXR precede the evaluated model reports. Publication chronology therefore cannot exclude prior exposure. The exact/perceptual screen concerns observable cross-dataset image reuse and the supervised experiment; it cannot certify that a foundation model has never seen an evaluation image. The study draws conclusions about the frozen configurations and annotated samples under examination. It does not estimate clinical utility, population-level TB prevalence or patient outcomes.
\FloatBarrier

\section{Results}
\label{sec:results}

\subsection{Model ranking does not summarize score reliability}

\begin{figure}[!htbp]
\centering
\begin{subfigure}[t]{.485\linewidth}
\centering\includegraphics[width=\linewidth]{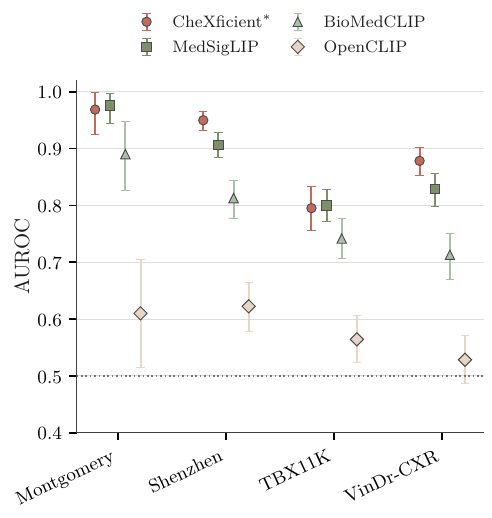}
\caption{Discrimination measured by AUROC.}
\end{subfigure}\hfill
\begin{subfigure}[t]{.485\linewidth}
\centering\includegraphics[width=\linewidth]{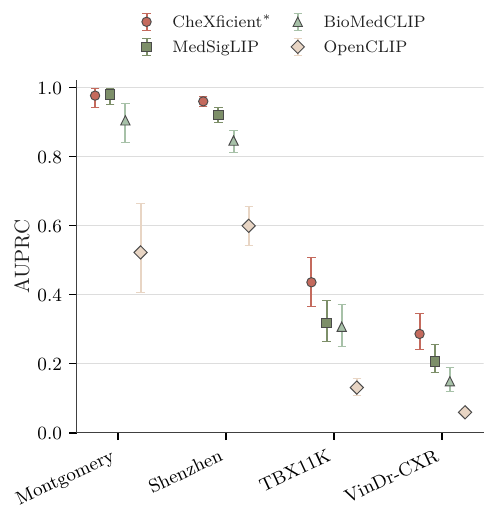}
\caption{Precision--recall performance at observed prevalence.}
\end{subfigure}
\caption{Primary clinical-prompt performance with 95\% image-level bootstrap intervals. Each model retains its checkpoint, processor and prompt rule across cohorts. AUPRC changes with class balance as well as score ordering. Intervals use ordinary image resampling in the original cohorts and within-label resampling in VinDr-CXR. $^{*}$The exposure marker applies to CheXficient on VinDr-CXR only: that cohort is a documented pretraining source.}
\label{fig:primary}
\end{figure}

The clinical-prompt comparison places CheXficient and MedSigLIP ahead of the other two models on the original primary cohorts (Figure~\ref{fig:primary}; Table~\ref{tab:primary}). MedSigLIP has the higher AUROC point estimate on Montgomery, 0.976 versus 0.969, and TBX11K validation, 0.800 versus 0.795. The paired MedSigLIP-minus-CheXficient differences are 0.007 (95\% CI $-0.007$ to 0.027) and 0.005 ($-0.023$ to 0.032), respectively. These intervals do not resolve a difference and do not establish equivalence. On Shenzhen, CheXficient reaches 0.950 versus MedSigLIP's 0.907, a difference of 0.043 (0.025 to 0.061).

Against all non-TB VinDr-CXR images, AUROC is 0.878 for CheXficient, 0.829 for MedSigLIP, 0.713 for BioMedCLIP and 0.528 for OpenCLIP. CheXficient's documented dataset exposure qualifies its apparent lead. The 0.878 estimate supports a within-dataset stress-test comparison, not a claim of independent external generalization.

Score reliability changes the comparison between the leading models. MedSigLIP has lower raw Brier score on Montgomery (0.101 versus 0.133) and Shenzhen (0.157 versus 0.197). CheXficient has lower Brier score on TBX11K validation (0.099 versus 0.171). It also has lower AURC there, 0.068 versus 0.204. Thus the small, unresolved AUROC difference on TBX11K validation coexists with substantial differences in score error and confidence-based deferral.

Table~\ref{tab:operating} separates within-cohort ROC operating points from selective prediction. These estimates assume access to the target ROC curve and cannot justify transporting a source threshold. The selective-risk results also depend on the fixed score threshold of 0.5. A model can rank TB images reasonably well yet assign score magnitudes that produce poor binary decisions or confidently retain errors.

\begin{table}[!htbp]
\centering
\caption{Primary clinical-prompt results across four evaluation cohorts.}
\label{tab:primary}
\small
\setlength{\tabcolsep}{4pt}
\renewcommand{\arraystretch}{1.12}
\begin{tabular}{lllrrrr}
\toprule
\headercell{Cohort} & \headercell{Model} & \headercell{AUROC [95\% CI]} & \headercell{AUPRC} & \headercell{Brier} & \headercell{ECE$_{15}$} & \headercell{AURC} \\
\midrule
Montgomery & CheXficient & 0.969 [0.925, 0.998] & 0.976 & 0.133 & 0.183 & 0.081 \\
\midrule
Montgomery & MedSigLIP & 0.976 [0.945, 0.998] & 0.979 & 0.101 & 0.181 & 0.039 \\
\midrule
Montgomery & BioMedCLIP & 0.890 [0.826, 0.947] & 0.905 & 0.173 & 0.200 & 0.108 \\
\midrule
Montgomery & OpenCLIP & 0.610 [0.515, 0.706] & 0.522 & 0.293 & 0.232 & 0.501 \\
\midrule
Shenzhen & CheXficient & 0.950 [0.932, 0.965] & 0.959 & 0.197 & 0.247 & 0.120 \\
\midrule
Shenzhen & MedSigLIP & 0.907 [0.884, 0.928] & 0.921 & 0.157 & 0.148 & 0.104 \\
\midrule
Shenzhen & BioMedCLIP & 0.813 [0.778, 0.843] & 0.846 & 0.299 & 0.315 & 0.224 \\
\midrule
Shenzhen & OpenCLIP & 0.623 [0.579, 0.664] & 0.599 & 0.271 & 0.164 & 0.411 \\
\midrule
TBX11K val. & CheXficient & 0.795 [0.755, 0.834] & 0.436 & 0.099 & 0.089 & 0.068 \\
\midrule
TBX11K val. & MedSigLIP & 0.800 [0.771, 0.829] & 0.318 & 0.171 & 0.257 & 0.204 \\
\midrule
TBX11K val. & BioMedCLIP & 0.742 [0.706, 0.778] & 0.307 & 0.148 & 0.131 & 0.141 \\
\midrule
TBX11K val. & OpenCLIP & 0.564 [0.523, 0.606] & 0.131 & 0.413 & 0.561 & 0.874 \\
\midrule
VinDr-CXR & CheXficient$^{*}$ & 0.878 [0.853, 0.902] & 0.286 & 0.075 & 0.106 & 0.030 \\
\midrule
VinDr-CXR & MedSigLIP & 0.829 [0.799, 0.856] & 0.207 & 0.146 & 0.282 & 0.169 \\
\midrule
VinDr-CXR & BioMedCLIP & 0.713 [0.669, 0.752] & 0.149 & 0.120 & 0.138 & 0.118 \\
\midrule
VinDr-CXR & OpenCLIP & 0.528 [0.486, 0.571] & 0.059 & 0.432 & 0.615 & 0.942 \\
\bottomrule
\end{tabular}
\par\smallskip
\begin{minipage}{0.98\linewidth}\footnotesize Lower Brier, ECE and AURC are preferable. AUPRC depends on prevalence. $^{*}$CheXficient has documented VinDr-CXR pretraining exposure. Original-cohort intervals use an ordinary image bootstrap; VinDr intervals use within-label resampling. These differences also apply to Figure~\ref{fig:primary}.\end{minipage}
\end{table}

\begin{table}[!htbp]
\centering
\caption{Within-cohort operating points and selective prediction under the clinical prompt.}
\label{tab:operating}
\small
\setlength{\tabcolsep}{4pt}
\renewcommand{\arraystretch}{1.12}
\begin{tabular}{llrrrr}
\toprule
\headercell{Cohort} & \headercell{Model} & \headercell{Se at 95\% Sp} & \headercell{Sp at 95\% Se} & \headercell{Risk at 80\%} & \headercell{Risk at 90\%} \\
\midrule
Montgomery & CheXficient & 0.948 & 0.938 & 0.117 & 0.152 \\
\midrule
Montgomery & MedSigLIP & 0.914 & 0.787 & 0.072 & 0.112 \\
\midrule
Montgomery & BioMedCLIP & 0.741 & 0.200 & 0.153 & 0.184 \\
\midrule
Montgomery & OpenCLIP & 0.138 & 0.075 & 0.577 & 0.568 \\
\midrule
Shenzhen & CheXficient & 0.839 & 0.644 & 0.208 & 0.247 \\
\midrule
Shenzhen & MedSigLIP & 0.670 & 0.521 & 0.181 & 0.206 \\
\midrule
Shenzhen & BioMedCLIP & 0.473 & 0.288 & 0.298 & 0.339 \\
\midrule
Shenzhen & OpenCLIP & 0.074 & 0.092 & 0.447 & 0.475 \\
\midrule
TBX11K val. & CheXficient & 0.420 & 0.120 & 0.078 & 0.098 \\
\midrule
TBX11K val. & MedSigLIP & 0.230 & 0.356 & 0.186 & 0.217 \\
\midrule
TBX11K val. & BioMedCLIP & 0.260 & 0.188 & 0.140 & 0.159 \\
\midrule
TBX11K val. & OpenCLIP & 0.045 & 0.052 & 0.882 & 0.888 \\
\midrule
VinDr-CXR & CheXficient$^{*}$ & 0.433 & 0.494 & 0.044 & 0.066 \\
\midrule
VinDr-CXR & MedSigLIP & 0.299 & 0.399 & 0.130 & 0.146 \\
\midrule
VinDr-CXR & BioMedCLIP & 0.262 & 0.173 & 0.106 & 0.120 \\
\midrule
VinDr-CXR & OpenCLIP & 0.055 & 0.092 & 0.941 & 0.944 \\
\bottomrule
\end{tabular}
\par\smallskip
\begin{minipage}{0.98\linewidth}\footnotesize Se: sensitivity; Sp: specificity. These operating points use each target cohort's ROC curve and are not transported source thresholds. Risk is the error fraction among retained images at the stated coverage, with a score threshold of 0.5. $^{*}$Documented VinDr-CXR exposure.\end{minipage}
\end{table}

\subsection{Prompt effects are model- and cohort-dependent}
\label{sec:prompt-results}

\begin{figure}[!htbp]
\centering
\begin{subfigure}[t]{.485\linewidth}
\centering\includegraphics[width=\linewidth]{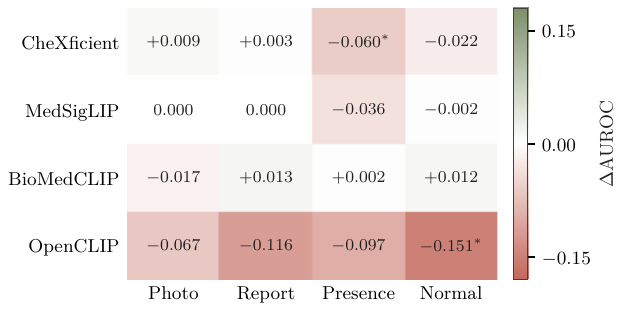}
\caption{Montgomery.}
\end{subfigure}\hfill
\begin{subfigure}[t]{.485\linewidth}
\centering\includegraphics[width=\linewidth]{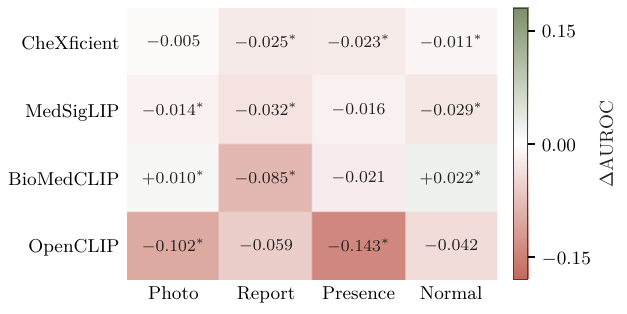}
\caption{Shenzhen.}
\end{subfigure}
\par\medskip
\begin{subfigure}[t]{.485\linewidth}
\centering\includegraphics[width=\linewidth]{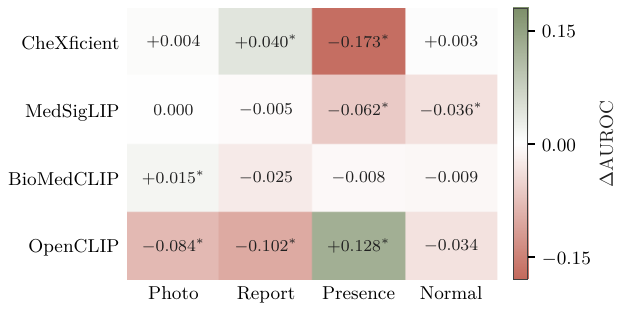}
\caption{TBX11K validation.}
\end{subfigure}\hfill
\begin{subfigure}[t]{.485\linewidth}
\centering\includegraphics[width=\linewidth]{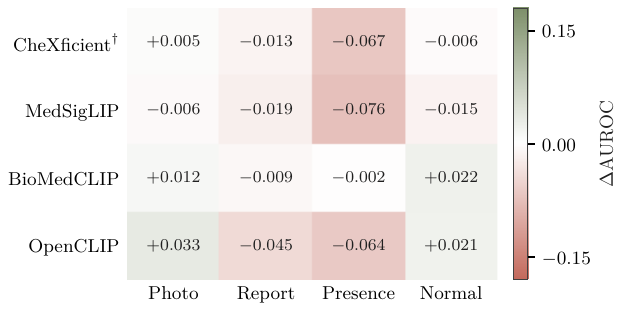}
\caption{VinDr-CXR test.}
\end{subfigure}
\caption{AUROC changes relative to the clinical prompt family. Each cell holds the checkpoint, images and labels fixed. Color scales are shared across panels. $^{*}$In (a)--(c), the paired DeLong test survives Benjamini--Hochberg correction across 48 comparisons. Panel (d) does not use that test family. $^{\dagger}$CheXficient has documented VinDr-CXR exposure. Photo, Report, Presence and Normal refer to the exact families in Table~\ref{tab:prompts}.}
\label{fig:prompts}
\end{figure}

Of the 48 primary AUROC contrasts, 21 meet the Benjamini--Hochberg criterion at 0.05; 24 meet the unadjusted 0.05 criterion. Figure~\ref{fig:prompts} shows the direction and size of each change. The family-wide correction covers the original three primary cohorts only. VinDr-CXR prompt differences in panel~(d) are descriptive extension results.

BioMedCLIP illustrates why one globally preferred prompt is difficult to infer. The report-style family produces 0.903 AUROC on Montgomery, compared with 0.890 for the clinical family, but 0.728 on Shenzhen, compared with 0.813. The normal-contrast family reaches 0.835 on Shenzhen. CheXficient's report-style AUROC rises from 0.795 to 0.835 on TBX11K validation, whereas its presence/absence family falls to 0.622. OpenCLIP exhibits the opposite pattern on that cohort: presence/absence rises from 0.564 to 0.692. These comparisons concern fixed alternatives, not prompts selected for deployment after observing the target results.

Table~\ref{tab:prompt-grid} gives all 100 model--cohort--prompt AUROC cells, including TBX11K training. The training split shows several patterns also present in validation: CheXficient favors report-style wording by point estimate, while OpenCLIP favors presence/absence. Agreement across labeled splits of one dataset is not independent clinical validation. The contrast between source and external performance remains the relevant test of broader reuse.

\begin{table}[!htbp]
\centering
\caption{Complete AUROC matrix for the five prompt families.}
\label{tab:prompt-grid}
\small
\setlength{\tabcolsep}{4pt}
\renewcommand{\arraystretch}{1.12}
\begin{tabular}{llrrrrr}
\toprule
\headercell{Cohort} & \headercell{Model} & \headercell{Clinical} & \headercell{Photo} & \headercell{Report} & \headercell{Presence} & \headercell{Normal} \\
\midrule
Montgomery & CheXficient & 0.969 & 0.978 & 0.972 & 0.909 & 0.947 \\
\midrule
Montgomery & MedSigLIP & 0.976 & 0.976 & 0.976 & 0.940 & 0.974 \\
\midrule
Montgomery & BioMedCLIP & 0.890 & 0.873 & 0.903 & 0.893 & 0.903 \\
\midrule
Montgomery & OpenCLIP & 0.610 & 0.543 & 0.494 & 0.514 & 0.459 \\
\midrule
Shenzhen & CheXficient & 0.950 & 0.945 & 0.925 & 0.927 & 0.939 \\
\midrule
Shenzhen & MedSigLIP & 0.907 & 0.892 & 0.874 & 0.890 & 0.878 \\
\midrule
Shenzhen & BioMedCLIP & 0.813 & 0.823 & 0.728 & 0.792 & 0.835 \\
\midrule
Shenzhen & OpenCLIP & 0.623 & 0.521 & 0.564 & 0.479 & 0.581 \\
\midrule
TBX11K val. & CheXficient & 0.795 & 0.800 & 0.835 & 0.622 & 0.798 \\
\midrule
TBX11K val. & MedSigLIP & 0.800 & 0.800 & 0.795 & 0.738 & 0.764 \\
\midrule
TBX11K val. & BioMedCLIP & 0.742 & 0.756 & 0.717 & 0.734 & 0.733 \\
\midrule
TBX11K val. & OpenCLIP & 0.564 & 0.481 & 0.462 & 0.692 & 0.530 \\
\midrule
VinDr-CXR & CheXficient$^{*}$ & 0.878 & 0.883 & 0.866 & 0.811 & 0.873 \\
\midrule
VinDr-CXR & MedSigLIP & 0.829 & 0.823 & 0.810 & 0.753 & 0.814 \\
\midrule
VinDr-CXR & BioMedCLIP & 0.713 & 0.725 & 0.704 & 0.712 & 0.735 \\
\midrule
VinDr-CXR & OpenCLIP & 0.528 & 0.561 & 0.483 & 0.465 & 0.549 \\
\midrule
TBX11K train & CheXficient & 0.816 & 0.806 & 0.856 & 0.650 & 0.820 \\
\midrule
TBX11K train & MedSigLIP & 0.820 & 0.822 & 0.815 & 0.748 & 0.782 \\
\midrule
TBX11K train & BioMedCLIP & 0.737 & 0.758 & 0.718 & 0.734 & 0.734 \\
\midrule
TBX11K train & OpenCLIP & 0.540 & 0.449 & 0.432 & 0.707 & 0.550 \\
\bottomrule
\end{tabular}
\par\smallskip
\begin{minipage}{0.98\linewidth}\footnotesize Clinical is the primary prompt family. Alternative maxima are descriptive, not selected deployment configurations. TBX11K training is a secondary replication cohort for zero-shot scoring. $^{*}$Documented VinDr-CXR exposure.\end{minipage}
\end{table}

Equal-weight averaging of the five family-level scores does not consistently remove sensitivity to the text specification (Table~\ref{tab:prompt-average}). AUROC increases in five of twelve original primary model--cohort cells and Brier score decreases in eight. These counts use point estimates, not statistical significance. For example, OpenCLIP's TBX11K AUROC change is $+0.024$, with an interval spanning $-0.007$ to $0.055$. Averaging can reduce score error while lowering discrimination, as in BioMedCLIP on Shenzhen. It is therefore not a general remedy for prompt-dependent conclusions.

A descriptive balanced decomposition of the original 60 primary model--cohort--prompt cells attributes 76.6\% of AUROC sum of squares to the model main effect, 10.9\% to cohort and 0.8\% to prompt alone (Table~\ref{tab:factorial}). The remaining share lies in interactions. For Brier score, the model--prompt interaction contributes 24.4\%; for ECE$_{15}$, it contributes 30.0\%. The small AUROC prompt main effect averages over directions that differ across models and cohorts. This decomposition has one observed value per cell and no independent residual replication. It summarizes this finite matrix, rather than supplying a causal variance attribution or an inferential ANOVA.

\begin{table}[!htbp]
\centering
\caption{Equal-weight averaging of all five prompt-family scores relative to the clinical prompt.}
\label{tab:prompt-average}
\small
\setlength{\tabcolsep}{4pt}
\renewcommand{\arraystretch}{1.12}
\begin{tabular}{llll}
\toprule
\headercell{Cohort} & \headercell{Model} & \headercell{$\Delta$AUROC [95\% CI]} & \headercell{$\Delta$Brier [95\% CI]} \\
\midrule
Montgomery & CheXficient & 0.005 [0.000, 0.014] & -0.017 [-0.024, -0.012] \\
\midrule
Montgomery & MedSigLIP & 0.003 [-0.002, 0.009] & 0.014 [0.011, 0.017] \\
\midrule
Montgomery & BioMedCLIP & 0.010 [-0.009, 0.032] & 0.002 [-0.003, 0.007] \\
\midrule
Montgomery & OpenCLIP & -0.076 [-0.152, 0.002] & -0.028 [-0.032, -0.024] \\
\midrule
Shenzhen & CheXficient & -0.007 [-0.011, -0.003] & -0.047 [-0.054, -0.041] \\
\midrule
Shenzhen & MedSigLIP & -0.014 [-0.018, -0.011] & 0.016 [0.015, 0.018] \\
\midrule
Shenzhen & BioMedCLIP & -0.028 [-0.043, -0.014] & -0.040 [-0.046, -0.035] \\
\midrule
Shenzhen & OpenCLIP & -0.026 [-0.057, 0.004] & -0.021 [-0.023, -0.019] \\
\midrule
TBX11K val. & CheXficient & 0.000 [-0.005, 0.006] & 0.047 [0.043, 0.051] \\
\midrule
TBX11K val. & MedSigLIP & -0.012 [-0.014, -0.009] & -0.009 [-0.011, -0.008] \\
\midrule
TBX11K val. & BioMedCLIP & -0.008 [-0.023, 0.007] & -0.008 [-0.011, -0.004] \\
\midrule
TBX11K val. & OpenCLIP & 0.024 [-0.007, 0.055] & -0.082 [-0.083, -0.080] \\
\bottomrule
\end{tabular}
\par\smallskip
\begin{minipage}{0.98\linewidth}\footnotesize Positive AUROC differences and negative Brier differences favor averaging. Intervals are paired descriptive intervals, without multiplicity correction.\end{minipage}
\end{table}

\begin{table}[!htbp]
\centering
\caption{Descriptive decomposition of the original balanced model--cohort--prompt matrix.}
\label{tab:factorial}
\small
\setlength{\tabcolsep}{4pt}
\renewcommand{\arraystretch}{1.12}
\begin{tabular}{lrrr}
\toprule
\headercell{Component} & \headercell{AUROC (\%)} & \headercell{Brier (\%)} & \headercell{ECE$_{15}$ (\%)} \\
\midrule
Model & 76.6 & 37.7 & 10.3 \\
\midrule
Cohort & 10.9 & 3.0 & 7.0 \\
\midrule
Prompt & 0.8 & 5.8 & 5.6 \\
\midrule
Model $\times$ cohort & 5.8 & 5.9 & 17.8 \\
\midrule
Model $\times$ prompt & 1.9 & 24.4 & 30.0 \\
\midrule
Cohort $\times$ prompt & 0.2 & 11.9 & 14.8 \\
\midrule
Model $\times$ cohort $\times$ prompt & 3.8 & 11.2 & 14.4 \\
\bottomrule
\end{tabular}
\par\smallskip
\begin{minipage}{0.98\linewidth}\footnotesize Values are shares of total sum of squares across four models, three primary cohorts and five prompt families (60 cells). There is one observed value per cell. The decomposition is descriptive, with no independent residual estimate or significance test; percentages may not sum to 100 after rounding.\end{minipage}
\end{table}

% \FloatBarrier

\subsection{Healthy controls conceal a harder discrimination problem}
\label{sec:negative-results}

\begin{figure}[!htbp]
\centering
\begin{subfigure}[t]{.485\linewidth}
\centering\includegraphics[width=\linewidth]{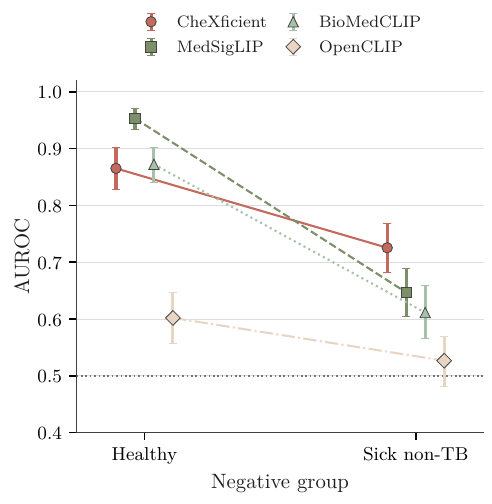}
\caption{TBX11K validation: 200 shared TB cases.}
\end{subfigure}\hfill
\begin{subfigure}[t]{.485\linewidth}
\centering\includegraphics[width=\linewidth]{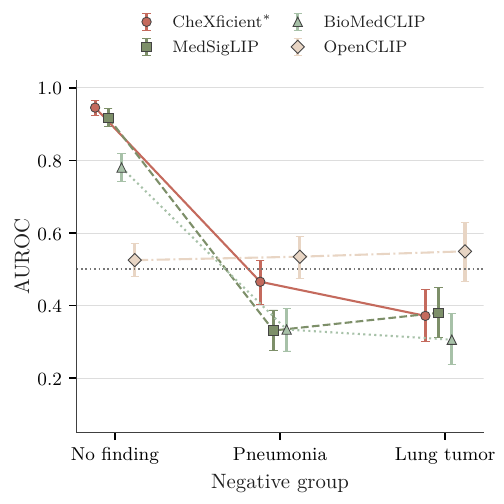}
\caption{VinDr-CXR: 164 shared TB cases.}
\end{subfigure}
\caption{Discrimination changes when the negative group changes. Lines connect the same model; error bars show marginal 95\% intervals. Panel (a) uses 800 controls per group. Panel (b) uses 2,051 no-finding, 210 pneumonia and 73 lung-tumor controls. The positive sample remains fixed within each panel. The dotted reference marks AUROC 0.5. $^{*}$CheXficient has documented VinDr-CXR pretraining exposure. Paired uncertainty for the negative-group difference is distinct from these marginal intervals.}
\label{fig:negative}
\end{figure}

Holding the TBX11K positive images fixed, replacing healthy with sick non-TB controls reduces clinical-prompt AUROC for every model (Figure~\ref{fig:negative}). The loss is 0.140 for CheXficient, 0.306 for MedSigLIP, 0.261 for BioMedCLIP and 0.075 for OpenCLIP. Every paired interval excludes zero, and all four Holm-adjusted permutation $p$ values are approximately $0.0004$ (Table~\ref{tab:negative-inference}). The equal numbers of healthy and sick controls remove prevalence and negative-sample-size differences from this comparison.

The decline is not limited to AUROC. MedSigLIP's AURC increases from 0.017 with healthy controls to 0.447 with sick controls. BioMedCLIP increases from 0.046 to 0.315, and CheXficient from 0.071 to 0.154. Confidence-based deferral therefore inherits the difficulty of the negative spectrum. A good healthy-control result does not demonstrate that a model's high-confidence TB predictions remain useful among patients with other abnormalities.

\begin{table}[!htbp]
\centering
\caption{Paired negative-spectrum inference on TBX11K validation.}
\label{tab:negative-inference}
\small
\setlength{\tabcolsep}{4pt}
\renewcommand{\arraystretch}{1.12}
\begin{tabular}{lrrlr}
\toprule
\headercell{Model} & \headercell{Healthy} & \headercell{Sick non-TB} & \headercell{AUROC loss [95\% CI]} & \headercell{Holm $p$} \\
\midrule
CheXficient & 0.865 & 0.725 & 0.140 [0.117, 0.163] & 0.0004 \\
\midrule
MedSigLIP & 0.953 & 0.647 & 0.306 [0.271, 0.344] & 0.0004 \\
\midrule
BioMedCLIP & 0.872 & 0.611 & 0.261 [0.231, 0.292] & 0.0004 \\
\midrule
OpenCLIP & 0.602 & 0.527 & 0.075 [0.048, 0.103] & 0.0004 \\
\bottomrule
\end{tabular}
\par\smallskip
\begin{minipage}{0.98\linewidth}\footnotesize Each contrast uses the same 200 TB images and 800 controls. Positive cases are resampled jointly across contrasts. Loss is healthy-control AUROC minus sick-control AUROC; Holm correction covers four models.\end{minipage}
\end{table}

VinDr-CXR makes the alternative diagnoses explicit (Table~\ref{tab:vindr-named}). CheXficient reaches 0.945 AUROC against no-finding controls, but 0.466 against pneumonia and 0.372 against lung tumor. MedSigLIP changes from 0.918 to 0.331 and 0.380; BioMedCLIP changes from 0.780 to 0.334 and 0.306. Each of the six medical-model named-minus-no-finding paired intervals excludes zero. OpenCLIP stays near 0.5 across the three spectra; small changes around weak discrimination do not establish clinically useful robustness.

All six medical-model point estimates are below 0.5 against the two named diseases, although CheXficient's pneumonia interval, 0.404--0.524, includes 0.5. Score direction is fixed before evaluation, so reversing it after observing these results would define a new, unvalidated classifier. Under the frozen prompt, several competing abnormalities receive higher TB-relative scores than many labeled TB cases.

\begin{table}[!htbp]
\centering
\caption{VinDr-CXR AUROC by named negative group under the clinical prompt.}
\label{tab:vindr-named}
\footnotesize
\setlength{\tabcolsep}{4pt}
\renewcommand{\arraystretch}{1.12}
\begin{tabular}{llll}
\toprule
\headercell{Model} & \headercell{No finding [95\% CI]} & \headercell{Pneumonia [95\% CI]} & \headercell{Lung tumor [95\% CI]} \\
\midrule
CheXficient$^{*}$ & 0.945 [0.924, 0.964] & 0.466 [0.404, 0.524] & 0.372 [0.302, 0.444] \\
\midrule
MedSigLIP & 0.918 [0.893, 0.942] & 0.331 [0.277, 0.388] & 0.380 [0.311, 0.451] \\
\midrule
BioMedCLIP & 0.780 [0.741, 0.819] & 0.334 [0.275, 0.392] & 0.306 [0.237, 0.380] \\
\midrule
OpenCLIP & 0.525 [0.481, 0.571] & 0.535 [0.474, 0.590] & 0.550 [0.468, 0.628] \\
\bottomrule
\end{tabular}
\par\smallskip
\begin{minipage}{0.98\linewidth}\footnotesize All three contrasts share the same 164 TB-positive images. Controls number 2,051, 210 and 73, respectively, after exclusion of TB-positive controls. $^{*}$Documented VinDr-CXR pretraining exposure. Values below 0.5 are point estimates; not every interval lies below 0.5.\end{minipage}
\end{table}

Table~\ref{tab:vindr-paired-spectrum} reports the paired differences directly. The shared positive resample accounts for the reuse of TB images across contrasts. It does not make the negative groups exchangeable clinical populations. These intervals quantify sampling variability within the specified subsets, rather than identify a causal effect of one diagnosis on the model.
\begin{table}[!htbp]
\centering
\caption{Paired VinDr-CXR AUROC differences relative to no-finding controls.}
\label{tab:vindr-paired-spectrum}
\small
\setlength{\tabcolsep}{4pt}
\renewcommand{\arraystretch}{1.12}
\begin{tabular}{lll}
\toprule
\headercell{Model} & \headercell{Pneumonia minus no finding} & \headercell{Lung tumor minus no finding} \\
\midrule
CheXficient$^{*}$ & -0.479 [-0.533, -0.426] & -0.573 [-0.640, -0.505] \\
\midrule
MedSigLIP & -0.586 [-0.634, -0.535] & -0.537 [-0.606, -0.470] \\
\midrule
BioMedCLIP & -0.446 [-0.492, -0.402] & -0.474 [-0.537, -0.408] \\
\midrule
OpenCLIP & 0.010 [-0.033, 0.051] & 0.024 [-0.048, 0.099] \\
\bottomrule
\end{tabular}
\par\smallskip
\begin{minipage}{0.98\linewidth}\footnotesize Values are differences [95\% CI]. Each bootstrap replicate reuses the positive resample and independently resamples the two negative groups. Intervals are descriptive. $^{*}$Documented VinDr-CXR exposure.\end{minipage}
\end{table}

\subsection{Class weighting changes the score-reliability comparison}

\begin{figure}[!htbp]
\centering
\begin{subfigure}[t]{.32\linewidth}
\centering\includegraphics[width=\linewidth]{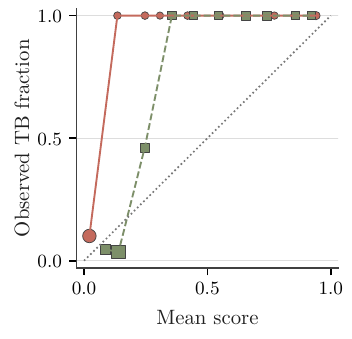}
\caption{Montgomery.}
\end{subfigure}\hfill
\begin{subfigure}[t]{.32\linewidth}
\centering\includegraphics[width=\linewidth]{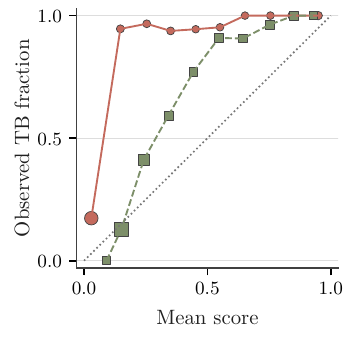}
\caption{Shenzhen.}
\end{subfigure}\hfill
\begin{subfigure}[t]{.32\linewidth}
\centering\includegraphics[width=\linewidth]{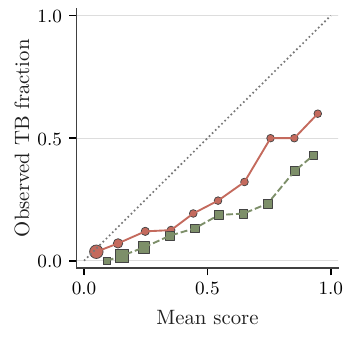}
\caption{TBX11K validation.}
\end{subfigure}
\caption{Raw-score reliability for CheXficient (coral circles, solid line) and MedSigLIP (olive squares, dashed line). Ten equal-width bins display mean score against observed TB fraction; larger markers indicate more images within a model--cohort panel. The diagonal denotes agreement. Empty bins are omitted. These plots do not use the 15-bin partition of ECE$_{15}$ and do not apply prevalence reweighting.}
\label{fig:reliability}
\end{figure}

The raw reliability diagrams for CheXficient and MedSigLIP show why their AUROC comparison is incomplete (Figure~\ref{fig:reliability}). Bin-wise observed TB fractions often differ from mean scores. These plots are descriptive: small bins, particularly in Montgomery, produce unstable fractions. Neither a visually favorable curve nor one small ECE estimate certifies probability calibration.

\begin{figure}[!htbp]
\centering
\begin{subfigure}[t]{.485\linewidth}
\centering\includegraphics[width=\linewidth]{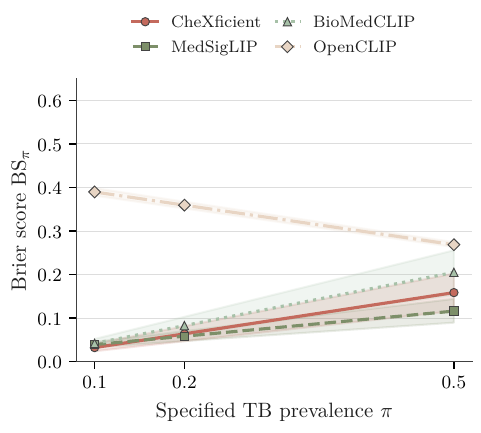}
\caption{Montgomery.}
\end{subfigure}\hfill
\begin{subfigure}[t]{.485\linewidth}
\centering\includegraphics[width=\linewidth]{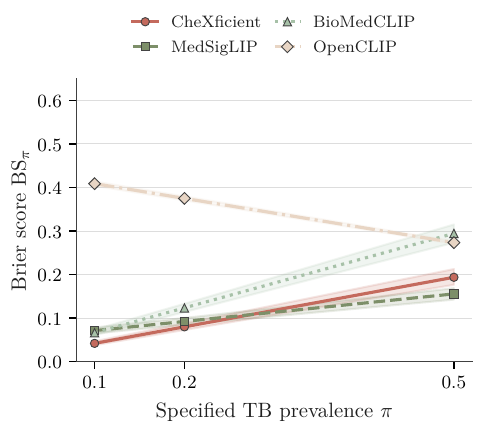}
\caption{Shenzhen.}
\end{subfigure}
\par\medskip
\begin{subfigure}[t]{.485\linewidth}
\centering\includegraphics[width=\linewidth]{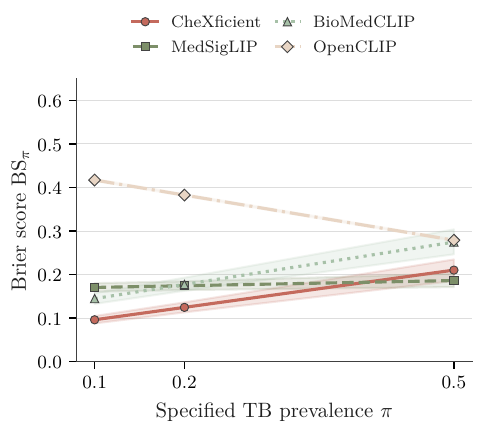}
\caption{TBX11K validation.}
\end{subfigure}\hfill
\begin{subfigure}[t]{.485\linewidth}
\centering\includegraphics[width=\linewidth]{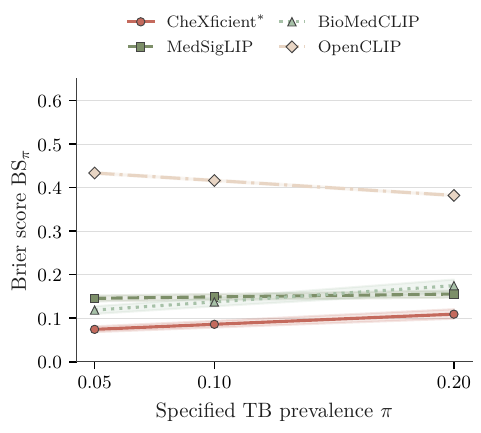}
\caption{VinDr-CXR, all non-TB controls.}
\end{subfigure}
\caption{Brier score under specified class weights. Lines connect evaluated prevalences; shaded regions show pointwise 95\% within-label bootstrap intervals. The vertical scale is shared. The VinDr panel has a different prevalence grid, explicitly marked on its horizontal axis. Scores and conditional samples remain unchanged. Brier score includes discrimination and calibration effects; a lower value does not by itself establish calibration. $^{*}$Documented VinDr-CXR exposure for CheXficient.}
\label{fig:calibration}
\end{figure}

Prevalence standardization changes the comparison between these models without changing their scores (Figure~\ref{fig:calibration}; Table~\ref{tab:calibration}). At 50\% specified TB prevalence, CheXficient's Brier score exceeds MedSigLIP's by 0.042 on Montgomery (95\% CI 0.013--0.074), 0.038 on Shenzhen (0.027--0.050) and 0.024 on TBX11K validation (0.006--0.041). At 10\%, the point-estimate ordering reverses in all three cohorts. The squared error assigned to positives and negatives differs between models, so changing their weights can change the preferred model even though AUROC is unchanged by this reweighting.

The VinDr-CXR extension uses its own stated prevalence grid rather than extrapolating a deployment prevalence from the dataset. Against no-finding controls, CheXficient's Brier score at 10\% TB prevalence is 0.039. Against pneumonia it is 0.419, and against lung tumor 0.525. Fixed prevalence therefore does not eliminate the negative-spectrum failure. Class balance and the distribution within the non-TB class affect different parts of the assessment.

\begin{table}[!htbp]
\centering
\caption{Brier scores standardized to specified TB prevalences.}
\label{tab:calibration}
\small
\setlength{\tabcolsep}{4pt}
\renewcommand{\arraystretch}{1.12}
\begin{tabular}{lllrrr}
\toprule
\headercell{Cohort} & \headercell{Model} & \headercell{Prevalences (\%)} & \headercell{First} & \headercell{Second} & \headercell{Third} \\
\midrule
Montgomery & CheXficient & 10/20/50 & 0.032 & 0.064 & 0.159 \\
\midrule
Montgomery & MedSigLIP & 10/20/50 & 0.039 & 0.058 & 0.116 \\
\midrule
Montgomery & BioMedCLIP & 10/20/50 & 0.042 & 0.083 & 0.205 \\
\midrule
Montgomery & OpenCLIP & 10/20/50 & 0.390 & 0.359 & 0.269 \\
\midrule
Shenzhen & CheXficient & 10/20/50 & 0.042 & 0.080 & 0.194 \\
\midrule
Shenzhen & MedSigLIP & 10/20/50 & 0.071 & 0.092 & 0.156 \\
\midrule
Shenzhen & BioMedCLIP & 10/20/50 & 0.066 & 0.123 & 0.294 \\
\midrule
Shenzhen & OpenCLIP & 10/20/50 & 0.409 & 0.375 & 0.273 \\
\midrule
TBX11K val. & CheXficient & 10/20/50 & 0.096 & 0.125 & 0.210 \\
\midrule
TBX11K val. & MedSigLIP & 10/20/50 & 0.170 & 0.174 & 0.186 \\
\midrule
TBX11K val. & BioMedCLIP & 10/20/50 & 0.145 & 0.177 & 0.275 \\
\midrule
TBX11K val. & OpenCLIP & 10/20/50 & 0.417 & 0.383 & 0.279 \\
\midrule
VinDr-CXR & CheXficient$^{*}$ & 5/10/20 & 0.074 & 0.086 & 0.109 \\
\midrule
VinDr-CXR & MedSigLIP & 5/10/20 & 0.146 & 0.149 & 0.155 \\
\midrule
VinDr-CXR & BioMedCLIP & 5/10/20 & 0.119 & 0.137 & 0.175 \\
\midrule
VinDr-CXR & OpenCLIP & 5/10/20 & 0.433 & 0.416 & 0.382 \\
\bottomrule
\end{tabular}
\par\smallskip
\begin{minipage}{0.98\linewidth}\footnotesize The final three columns follow the prevalence order in each row. Standardization changes class weights, not the conditional score distributions. It does not recalibrate the model. Figure~\ref{fig:calibration} shows 95\% intervals. $^{*}$Documented VinDr-CXR exposure.\end{minipage}
\end{table}

Table~\ref{tab:weighted-reliability} adds weighted ECE$_{15}$ and log loss at each block's primary prevalence. These endpoints describe different aspects of score behavior: ECE summarizes binned agreement, while log loss strongly penalizes confident errors. Neither changes the evaluated predictions. Reporting them together reduces the risk of treating a favorable Brier score as a complete calibration assessment.
\begin{table}[!htbp]
\centering
\caption{Additional standardized score-reliability endpoints at each block's primary prevalence.}
\label{tab:weighted-reliability}
\footnotesize
\setlength{\tabcolsep}{4pt}
\renewcommand{\arraystretch}{1.12}
\begin{tabular}{lll ll}
\toprule
\headercell{Cohort} & \headercell{Model} & \headercell{$\pi$} & \headercell{ECE$_{15}$ [95\% CI]} & \headercell{Log loss [95\% CI]} \\
\midrule
Montgomery & CheXficient & 0.50 & 0.221 [0.180, 0.264] & 0.517 [0.372, 0.681] \\
\midrule
Montgomery & MedSigLIP & 0.50 & 0.198 [0.157, 0.243] & 0.376 [0.313, 0.446] \\
\midrule
Montgomery & BioMedCLIP & 0.50 & 0.241 [0.190, 0.291] & 0.801 [0.583, 1.049] \\
\midrule
Montgomery & OpenCLIP & 0.50 & 0.153 [0.148, 0.193] & 0.732 [0.720, 0.745] \\
\midrule
Shenzhen & CheXficient & 0.50 & 0.243 [0.226, 0.261] & 0.615 [0.555, 0.679] \\
\midrule
Shenzhen & MedSigLIP & 0.50 & 0.146 [0.127, 0.171] & 0.476 [0.446, 0.508] \\
\midrule
Shenzhen & BioMedCLIP & 0.50 & 0.310 [0.291, 0.332] & 1.078 [0.987, 1.170] \\
\midrule
Shenzhen & OpenCLIP & 0.50 & 0.172 [0.169, 0.175] & 0.743 [0.736, 0.750] \\
\midrule
TBX11K val. & CheXficient & 0.50 & 0.169 [0.144, 0.196] & 0.679 [0.599, 0.765] \\
\midrule
TBX11K val. & MedSigLIP & 0.50 & 0.072 [0.058, 0.107] & 0.555 [0.520, 0.593] \\
\midrule
TBX11K val. & BioMedCLIP & 0.50 & 0.244 [0.215, 0.282] & 0.950 [0.846, 1.060] \\
\midrule
TBX11K val. & OpenCLIP & 0.50 & 0.176 [0.173, 0.182] & 0.755 [0.751, 0.760] \\
\midrule
VinDr-CXR & CheXficient$^{*}$ & 0.10 & 0.080 [0.072, 0.090] & 0.287 [0.265, 0.311] \\
\midrule
VinDr-CXR & MedSigLIP & 0.10 & 0.249 [0.242, 0.257] & 0.479 [0.461, 0.496] \\
\midrule
VinDr-CXR & BioMedCLIP & 0.10 & 0.133 [0.121, 0.148] & 0.522 [0.478, 0.564] \\
\midrule
VinDr-CXR & OpenCLIP & 0.10 & 0.570 [0.568, 0.572] & 1.046 [1.041, 1.051] \\
\bottomrule
\end{tabular}
\par\smallskip
\begin{minipage}{0.98\linewidth}\footnotesize Lower values are preferable. The different primary prevalence settings are stated in each row and prevent direct interpretation as one common-prevalence comparison. $^{*}$Documented VinDr-CXR exposure.\end{minipage}
\end{table}

% \FloatBarrier

\subsection{Source sensitivity constraints rarely survive unchanged}
\label{sec:threshold-results}

\begin{figure}[!htbp]
\centering
\begin{subfigure}[t]{.485\linewidth}
\centering\includegraphics[width=\linewidth]{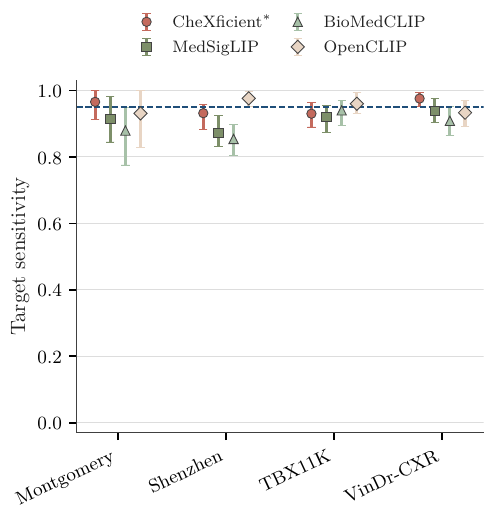}
\caption{Target sensitivity at the unchanged source threshold.}
\end{subfigure}\hfill
\begin{subfigure}[t]{.485\linewidth}
\centering\includegraphics[width=\linewidth]{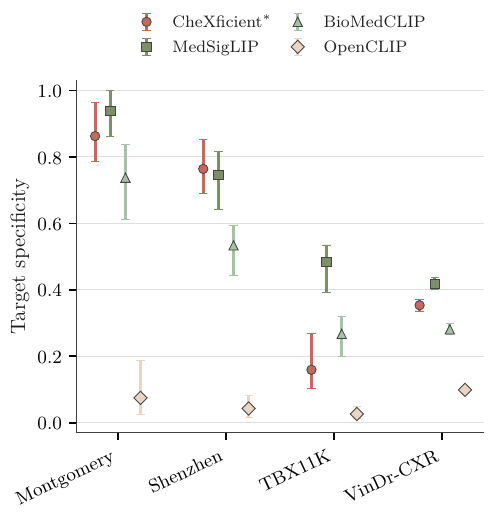}
\caption{Target specificity at the same threshold.}
\end{subfigure}
\caption{Transport of a threshold chosen for 95\% sensitivity on TBX11K training. The dashed line in (a) marks the source constraint. Error bars show 95\% intervals. Original-target intervals incorporate source-threshold re-estimation; VinDr intervals condition on the frozen source threshold and quantify target sampling uncertainty only. $^{*}$CheXficient's VinDr estimate has documented dataset exposure. Retention counts use point estimates, not confidence bounds.}
\label{fig:transport}
\end{figure}

The source thresholds vary substantially across models, despite targeting the same 95\% sensitivity criterion (Table~\ref{tab:source-thresholds}). Applying them unchanged retains at least 95\% target sensitivity in only four of sixteen transports (Figure~\ref{fig:transport}; Table~\ref{tab:threshold}). CheXficient retains the criterion on Montgomery and VinDr-CXR. OpenCLIP retains it on Shenzhen and TBX11K validation. BioMedCLIP and MedSigLIP do not retain it on any of the four targets under this source choice.

Retention alone can be misleading. OpenCLIP's retained constraints accompany specificity of approximately 0.043 on Shenzhen and 0.026 on TBX11K validation. These operating rules label nearly all non-TB images as positive. CheXficient reaches VinDr-CXR sensitivity of 0.976 and specificity of 0.354 against all non-TB images. Specificity falls to 0.005 against pneumonia and 0.000 against lung tumor, corresponding to one correctly rejected pneumonia control out of 210 and none of 73 tumor controls. Its documented exposure does not prevent this named-spectrum operating failure.

\begin{table}[!htbp]
\centering
\caption{Operating thresholds estimated from all 6,600 TBX11K training images.}
\label{tab:source-thresholds}
\small
\setlength{\tabcolsep}{4pt}
\renewcommand{\arraystretch}{1.12}
\begin{tabular}{llrrr}
\toprule
\headercell{Model} & \headercell{Source constraint} & \headercell{Threshold $\tau$} & \headercell{Source Se} & \headercell{Source Sp} \\
\midrule
CheXficient & 95\% sensitivity & 0.032724 & 0.950 & 0.172 \\
\midrule
CheXficient & 95\% specificity & 0.685756 & 0.392 & 0.950 \\
\midrule
MedSigLIP & 95\% sensitivity & 0.225860 & 0.950 & 0.478 \\
\midrule
MedSigLIP & 95\% specificity & 0.830078 & 0.272 & 0.950 \\
\midrule
BioMedCLIP & 95\% sensitivity & 0.017594 & 0.950 & 0.279 \\
\midrule
BioMedCLIP & 95\% specificity & 0.921753 & 0.212 & 0.950 \\
\midrule
OpenCLIP & 95\% sensitivity & 0.606691 & 0.950 & 0.028 \\
\midrule
OpenCLIP & 95\% specificity & 0.725010 & 0.068 & 0.950 \\
\bottomrule
\end{tabular}
\end{table}

\begin{table}[!htbp]
\centering
\caption{Transport of the TBX11K-training threshold chosen for 95\% sensitivity.}
\label{tab:threshold}
\footnotesize
\setlength{\tabcolsep}{4pt}
\renewcommand{\arraystretch}{1.12}
\begin{tabular}{llllc}
\toprule
\headercell{Target} & \headercell{Model} & \headercell{Sensitivity [95\% CI]} & \headercell{Specificity [95\% CI]} & \headercell{Retains} \\
\midrule
Montgomery & CheXficient & 0.966 [0.914, 1.000] & 0.863 [0.787, 0.963] & Yes \\
\midrule
Montgomery & MedSigLIP & 0.914 [0.845, 0.983] & 0.938 [0.863, 1.000] & No \\
\midrule
Montgomery & BioMedCLIP & 0.879 [0.776, 0.948] & 0.738 [0.613, 0.838] & No \\
\midrule
Montgomery & OpenCLIP & 0.931 [0.828, 1.000] & 0.075 [0.025, 0.188] & No \\
\midrule
Shenzhen & CheXficient & 0.932 [0.884, 0.958] & 0.764 [0.690, 0.853] & No \\
\midrule
Shenzhen & MedSigLIP & 0.872 [0.830, 0.926] & 0.745 [0.641, 0.816] & No \\
\midrule
Shenzhen & BioMedCLIP & 0.854 [0.804, 0.899] & 0.534 [0.445, 0.595] & No \\
\midrule
Shenzhen & OpenCLIP & 0.976 [0.952, 0.997] & 0.043 [0.015, 0.083] & Yes \\
\midrule
TBX11K val. & CheXficient & 0.930 [0.890, 0.965] & 0.159 [0.104, 0.268] & No \\
\midrule
TBX11K val. & MedSigLIP & 0.920 [0.875, 0.955] & 0.484 [0.393, 0.535] & No \\
\midrule
TBX11K val. & BioMedCLIP & 0.940 [0.895, 0.970] & 0.268 [0.200, 0.321] & No \\
\midrule
TBX11K val. & OpenCLIP & 0.960 [0.930, 0.995] & 0.026 [0.012, 0.050] & Yes \\
\midrule
VinDr-CXR & CheXficient$^{*}$ & 0.976 [0.951, 0.994] & 0.354 [0.336, 0.372] & Yes \\
\midrule
VinDr-CXR & MedSigLIP & 0.939 [0.902, 0.976] & 0.419 [0.401, 0.437] & No \\
\midrule
VinDr-CXR & BioMedCLIP & 0.909 [0.866, 0.951] & 0.281 [0.265, 0.298] & No \\
\midrule
VinDr-CXR & OpenCLIP & 0.933 [0.890, 0.970] & 0.099 [0.088, 0.110] & No \\
\bottomrule
\end{tabular}
\par\smallskip
\begin{minipage}{0.98\linewidth}\footnotesize Retains refers to target point-estimate sensitivity $\geq0.95$, not a lower confidence bound. Original-target intervals resample both source and target and re-estimate the threshold. VinDr intervals condition on the fixed source threshold and resample only target images. $^{*}$Documented VinDr-CXR exposure.\end{minipage}
\end{table}

Changing the source does not remove dependence on the operating specification. With Shenzhen as source, four of eight transports retain 95\% sensitivity (Table~\ref{tab:threshold-shenzhen}). This result does not validate one source as universally preferable; it demonstrates that the source population belongs in the claim. The same target can receive a different binary classifier solely because its threshold comes from a different dataset.

The 95\%-specificity source rule gives a complementary picture (Table~\ref{tab:threshold-specificity}). Eleven of sixteen target point estimates retain that specificity constraint, but sensitivities range from 0.000 to 0.451. A constraint may therefore travel more often while the resulting classifier misses most TB-labeled cases. These values reinforce the need to state the error constraint and report its complementary operating characteristic.

\begin{table}[!htbp]
\centering
\caption{Sensitivity-threshold transport with Shenzhen as the alternative source.}
\label{tab:threshold-shenzhen}
\small
\setlength{\tabcolsep}{4pt}
\renewcommand{\arraystretch}{1.12}
\begin{tabular}{llrlr}
\toprule
\headercell{Target} & \headercell{Model} & \headercell{Threshold} & \headercell{Sensitivity [95\% CI]} & \headercell{Specificity} \\
\midrule
Montgomery & BioMedCLIP & 0.00694 & 0.914 [0.828, 0.983] & 0.438 \\
\midrule
TBX11K val. & BioMedCLIP & 0.00694 & 0.985 [0.965, 1.000] & 0.095 \\
\midrule
Montgomery & CheXficient & 0.02426 & 0.966 [0.914, 1.000] & 0.812 \\
\midrule
TBX11K val. & CheXficient & 0.02426 & 0.955 [0.910, 0.985] & 0.072 \\
\midrule
Montgomery & MedSigLIP & 0.17374 & 0.948 [0.897, 1.000] & 0.800 \\
\midrule
TBX11K val. & MedSigLIP & 0.17374 & 0.975 [0.930, 0.995] & 0.311 \\
\midrule
Montgomery & OpenCLIP & 0.61784 & 0.828 [0.724, 0.948] & 0.175 \\
\midrule
TBX11K val. & OpenCLIP & 0.61784 & 0.940 [0.895, 0.975] & 0.059 \\
\bottomrule
\end{tabular}
\par\smallskip
\begin{minipage}{0.98\linewidth}\footnotesize Thresholds target 95\% source sensitivity. Each bootstrap replicate re-estimates the source threshold and resamples target images independently.\end{minipage}
\end{table}

\begin{table}[!htbp]
\centering
\caption{Secondary transport of the TBX11K-training threshold chosen for 95\% specificity.}
\label{tab:threshold-specificity}
\small
\setlength{\tabcolsep}{4pt}
\renewcommand{\arraystretch}{1.12}
\begin{tabular}{llrrc}
\toprule
\headercell{Target} & \headercell{Model} & \headercell{Sensitivity} & \headercell{Specificity} & \headercell{Retains specificity} \\
\midrule
Montgomery & CheXficient & 0.379 & 1.000 & Yes \\
\midrule
Montgomery & MedSigLIP & 0.241 & 1.000 & Yes \\
\midrule
Montgomery & BioMedCLIP & 0.293 & 1.000 & Yes \\
\midrule
Montgomery & OpenCLIP & 0.000 & 0.988 & Yes \\
\midrule
Shenzhen & CheXficient & 0.369 & 1.000 & Yes \\
\midrule
Shenzhen & MedSigLIP & 0.190 & 1.000 & Yes \\
\midrule
Shenzhen & BioMedCLIP & 0.167 & 1.000 & Yes \\
\midrule
Shenzhen & OpenCLIP & 0.074 & 0.951 & Yes \\
\midrule
TBX11K val. & CheXficient & 0.390 & 0.954 & Yes \\
\midrule
TBX11K val. & MedSigLIP & 0.245 & 0.943 & No \\
\midrule
TBX11K val. & BioMedCLIP & 0.260 & 0.949 & No \\
\midrule
TBX11K val. & OpenCLIP & 0.050 & 0.947 & No \\
\midrule
VinDr-CXR & CheXficient$^{*}$ & 0.451 & 0.949 & No \\
\midrule
VinDr-CXR & MedSigLIP & 0.244 & 0.961 & Yes \\
\midrule
VinDr-CXR & BioMedCLIP & 0.238 & 0.953 & Yes \\
\midrule
VinDr-CXR & OpenCLIP & 0.128 & 0.887 & No \\
\bottomrule
\end{tabular}
\par\smallskip
\begin{minipage}{0.98\linewidth}\footnotesize The threshold remains unchanged at the target. Retention uses the target point estimate. $^{*}$Documented VinDr-CXR exposure.\end{minipage}
\end{table}

% \FloatBarrier

\subsection{Near-perfect supervised validation does not resolve transfer}
\label{sec:supervised-results}

\begin{figure}[!htbp]
\centering
\begin{subfigure}[t]{.485\linewidth}
\centering\includegraphics[width=\linewidth]{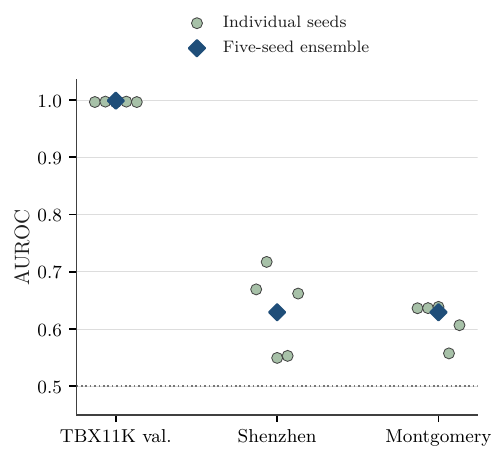}
\caption{Source-trained performance before candidate exclusion.}
\end{subfigure}\hfill
\begin{subfigure}[t]{.485\linewidth}
\centering\includegraphics[width=\linewidth]{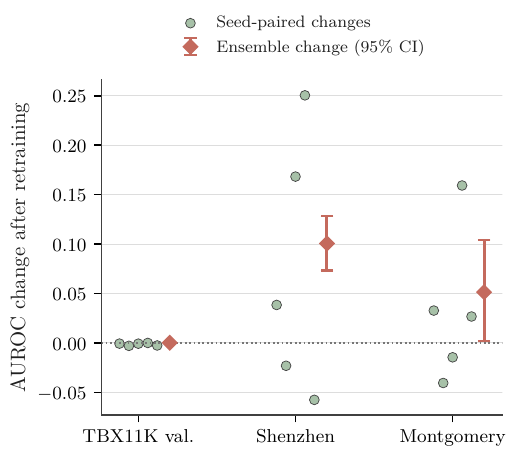}
\caption{Retraining effects on identical retained targets.}
\end{subfigure}
\caption{Supervised transfer and overlap-candidate sensitivity. Panel (a) shows each training seed and the mean-score ensemble. Panel (b) distinguishes seed-paired AUROC changes from the ensemble-level paired image-bootstrap interval. Positive changes favor retraining after exclusion. The ensemble interval conditions on two fitted ensembles and does not incorporate repeated training uncertainty.}
\label{fig:supervised}
\end{figure}

The supervised ResNet-50 reaches mean TBX11K validation AUROC of 0.997 across five seeds, with between-seed SD of approximately 0.001. The mean-score ensemble reaches 0.999. The same ensemble reaches only 0.629 on Shenzhen and 0.629 on Montgomery (Table~\ref{tab:supervised}; Figure~\ref{fig:supervised}). No target labels enter its checkpoint selection. The result therefore documents a source-to-external gap under one conventional supervised protocol, rather than an optimization trade-off chosen to favor the external cohorts.

\begin{table}[!htbp]
\centering
\caption{Supervised ResNet-50 source training and unchanged external evaluation.}
\label{tab:supervised}
\footnotesize
\setlength{\tabcolsep}{4pt}
\renewcommand{\arraystretch}{1.12}
\begin{tabular}{llrrrr}
\toprule
\headercell{Cohort} & \headercell{Seed AUROC mean $\pm$ SD} & \headercell{Ensemble AUROC} & \headercell{AUPRC} & \headercell{Brier} & \headercell{AURC} \\
\midrule
TBX11K val. & 0.997 $\pm$ 0.001 & 0.999 & 0.993 & 0.008 & 0.000 \\
\midrule
Shenzhen & 0.630 $\pm$ 0.075 & 0.629 & 0.691 & 0.402 & 0.346 \\
\midrule
Montgomery & 0.615 $\pm$ 0.035 & 0.629 & 0.583 & 0.269 & 0.323 \\
\bottomrule
\end{tabular}
\par\smallskip
\begin{minipage}{0.98\linewidth}\footnotesize SD summarizes five training seeds on a shared split, not a confidence interval. Remaining columns describe the mean-score ensemble. Source selection uses a disjoint 1,320-image subset of TBX11K training; TBX11K validation remains outside checkpoint selection.\end{minipage}
\end{table}

The cross-dataset screen identifies no byte-identical pairs in the screened material, but 249 perceptual candidates involve 313 evaluation images. Conservative exclusion removes 166 TBX11K training images, 37 validation images, 109 Shenzhen images and one Montgomery image (Table~\ref{tab:overlap-counts}). The retained optimization and selection subsets contain 5,151 and 1,283 images. These counts describe risk-sensitive exclusions, not confirmed duplicate counts.

\begin{table}[!htbp]
\centering
\caption{Conservative removal of all images implicated by the perceptual candidate screen.}
\label{tab:overlap-counts}
\small
\setlength{\tabcolsep}{4pt}
\renewcommand{\arraystretch}{1.12}
\begin{tabular}{lrrrrr}
\toprule
\headercell{Cohort} & \headercell{Original} & \headercell{Excluded} & \headercell{Retained} & \headercell{TB retained} & \headercell{Non-TB retained} \\
\midrule
Montgomery & 138 & 1 & 137 & 58 & 79 \\
\midrule
Shenzhen & 662 & 109 & 553 & 284 & 269 \\
\midrule
TBX11K train & 6600 & 166 & 6434 & 589 & 5845 \\
\midrule
TBX11K val. & 1800 & 37 & 1763 & 197 & 1566 \\
\bottomrule
\end{tabular}
\par\smallskip
\begin{minipage}{0.98\linewidth}\footnotesize The screen produces 249 candidate pairs involving 313 images. It does not adjudicate duplicates. Retained TBX11K training includes 5,151 optimization and 1,283 selection images, with their original split roles preserved.\end{minipage}
\end{table}

After retraining, the five-seed ensemble reaches 0.999 on retained TBX11K validation, 0.733 on retained Shenzhen and 0.678 on retained Montgomery. The appropriate paired comparison uses the original ensemble on those same retained targets, where its AUROCs are 0.999, 0.633 and 0.627 (Table~\ref{tab:overlap-effects}). The retraining differences are approximately 0.0002 ($-0.0003$ to 0.0008), 0.101 (0.073--0.128) and 0.051 (0.002--0.104), respectively. The intervals condition on the two trained ensembles.

The seed-level evidence is less uniform. Seed-paired retraining changes average 0.075 with SD 0.130 on Shenzhen and 0.033 with SD 0.077 on Montgomery. Two of five seeds decline on each external cohort. The positive ensemble effect consequently does not imply that every training run benefits from exclusion. The remaining source-to-external ensemble gaps are approximately 0.266 and 0.321 AUROC. Removing all candidates does not close them, and the experiment cannot attribute the improvement to elimination of verified image leakage.

\begin{table}[!htbp]
\centering
\caption{Retraining after candidate exclusion, evaluated on identical retained target images.}
\label{tab:overlap-effects}
\footnotesize
\setlength{\tabcolsep}{4pt}
\renewcommand{\arraystretch}{1.12}
\begin{tabular}{lrrll}
\toprule
\headercell{Cohort} & \headercell{Original retained} & \headercell{Retrained} & \headercell{Ensemble $\Delta$ [95\% CI]} & \headercell{Seed $\Delta$ mean $\pm$ SD} \\
\midrule
TBX11K val. & 0.9989 & 0.9991 & 0.0002 [-0.0003, 0.0008] & -0.001 $\pm$ 0.001 \\
\midrule
Shenzhen & 0.6329 & 0.7335 & 0.1006 [0.0732, 0.1282] & 0.075 $\pm$ 0.130 \\
\midrule
Montgomery & 0.6266 & 0.6779 & 0.0513 [0.0024, 0.1041] & 0.033 $\pm$ 0.077 \\
\bottomrule
\end{tabular}
\par\smallskip
\begin{minipage}{0.98\linewidth}\footnotesize The paired interval conditions on the two fitted five-seed ensembles and resamples retained images within label. Seed-paired changes summarize a different uncertainty source: training variability. Neither estimate establishes a causal effect of confirmed duplicate removal.\end{minipage}
\end{table}

\FloatBarrier

\section{Discussion}
\label{sec:discussion}

\subsection{Portability depends on the claim under evaluation}

The study connects several ways in which a favorable evaluation can support an overly broad conclusion. Cohort transfer changes the apparent leading model. Prompt specification changes the classifier on the same images. Negative-spectrum replacement changes the difficulty of the comparison even when the positive cases stay fixed. Prevalence weighting changes score-error comparisons, while an unchanged source threshold can lose its intended sensitivity. These are related but non-interchangeable failure modes.

Distribution shift, prompt sensitivity and calibration error are established phenomena. This study links them to distinct portability claims within one screening-oriented task. A model-ranking claim, a score-reliability claim and an operating-constraint claim can disagree for the same checkpoint. The paired prompt and TBX11K negative-spectrum comparisons isolate dependencies that a conventional cross-dataset leaderboard leaves unresolved.

This framing complements recent calls for contextual, clinically meaningful evaluation \citep{xian2025robustness,bielick2026benchmarks}. Equation~\ref{eq:specification} makes the assumptions explicit enough to test. It is not a new certification criterion: adequacy still depends on intended use, acceptable errors and local evidence. The result supports narrower, checkable claims rather than a binary designation of a model as portable or nonportable.

\subsection{The negative class defines the clinical question}

TB versus normal is a different task from TB versus another radiographic disease. The TBX11K comparison holds both the positive sample and class balance fixed, yet AUROC drops for every model when healthy controls are replaced by sick controls. VinDr-CXR then identifies concrete competing diagnoses. Pneumonia and lung tumor often receive higher TB-relative scores than TB-labeled images under the primary prompt.

These results are consistent with a score that responds partly to general abnormality or to visual features shared by diseases. They do not establish the underlying mechanism. We do not localize the model's evidence, adjudicate lesions, or have microbiological confirmation for every TB label. Co-occurring findings and labeling conventions may contribute. The supported conclusion is that healthy-control discrimination substantially overstates named-disease discrimination in these configurations.

The finding also qualifies the interpretation of a high aggregate VinDr-CXR AUROC. More than two thousand no-finding images contribute to the all-non-TB denominator, whereas pneumonia and lung-tumor controls are much less numerous. The aggregate can remain favorable while a clinically important comparison is poor. This is an instance of the broader hidden-stratification problem \citep{oakdenrayner2020hidden}, measured here through paired positive samples rather than inferred from an aggregate score alone.

\subsection{A prompt is part of the evaluated system}

The text specifies what counts as positive and what the model should contrast it with. A normal-radiograph negative can encourage a different ranking from a negative that allows abnormality but excludes TB. The clinical ensemble also differs structurally from the single-pair alternatives. Prompt-family effects therefore combine linguistic and semantic changes with the number of prototypes. They should not all be described as fragility to a harmless wording edit.

A zero-shot result therefore needs exact positive and negative text, aggregation rules and a selection policy. Reporting only a checkpoint name omits part of the classifier. Averaging more prompt families may help some scores, but the secondary average improves AUROC in fewer than half of the original primary cells by point estimate. A larger ensemble does not consistently stabilize the conclusions in this audit.

\subsection{Ranking, probabilities and screening rules answer different questions}

AUROC describes ordering, not numerical probability accuracy or the consequences of a chosen threshold. Two-class similarity scores can discriminate without being calibrated TB probabilities. The model's learned scale, prompt prototypes and negative text all affect their magnitude. Score reliability is therefore a property of the evaluated scoring pipeline and requires separate evidence.

Prevalence standardization clarifies one source of disagreement between reliability comparisons. It makes the relative weight of positive and negative errors explicit while retaining their observed conditional distributions. The reversal between CheXficient and MedSigLIP across specified prevalences follows from these different error profiles. It does not prove either model is well calibrated, and it cannot account for new conditional distributions in a future clinic. Calibration assessment and updating require their own evidence \citep{vancalster2019calibration}.

Threshold transport asks an operationally different question: does the same score cutoff retain its intended constraint? Four of sixteen source-sensitivity rules do so by point estimate, and some retained constraints coincide with extremely low specificity. The converse analysis shows that specificity can remain high while sensitivity becomes poor. Reporting both characteristics prevents a nominal constraint from becoming a misleading summary of the rule's usefulness.

The study does not evaluate a locally recalibrated or adapted system. Such methods may improve target performance, but they would define a new specification with a local training sample, selection policy and validation requirement. A threshold tuned retrospectively on the target cannot provide evidence for unchanged threshold transport.

\subsection{Supervision and exposure require separate qualifications}

The supervised experiment shows that excellent source validation is insufficient evidence of external usefulness. ResNet-50 performs nearly perfectly on the held-out source split but much less well on the two external NLM cohorts. It is a conventional reference, not evidence that supervised learning cannot generalize. A different training corpus, architecture or adaptation protocol may change the result.

The candidate-exclusion analysis strengthens the interpretation of the observed gap without resolving all leakage questions. By comparing original and retrained ensembles on identical retained targets, it avoids attributing a target-composition change to training. The external gains remain variable across seeds and leave substantial gaps. Unadjudicated perceptual candidates, limited reference coverage and possible within-dataset dependencies prevent a causal claim about duplicate removal.

Foundation-model exposure requires a separate qualification. CheXficient's documented VinDr-CXR use means that its result cannot serve as independent external validation. It remains informative as a within-dataset test of whether changing the negative spectrum or preserving a source threshold changes the conclusion. For the other checkpoints, undocumented membership remains unresolved; the available evidence supports neither confirmed independence nor confirmed contamination.

\subsection{Implications for evaluation practice}

A useful portability claim should name the model implementation, prompt, target population, controls, score interpretation and decision rule. It should also state what type of conclusion transfers. Table~\ref{tab:reporting} translates the observed failure modes into concrete reporting questions. The table is not a validated checklist or a substitute for prospective clinical evaluation.

\begin{table}[!htbp]
\centering
\caption{Information needed to interpret a portability claim in this task.}
\label{tab:reporting}
\small
\setlength{\tabcolsep}{4pt}
\begin{tabular}{>{\raggedright\arraybackslash}p{.19\linewidth}>{\raggedright\arraybackslash}p{.36\linewidth}>{\raggedright\arraybackslash}p{.37\linewidth}}
\toprule
\headercell{Claim component} & \headercell{Question to state explicitly} & \headercell{Evidence from this audit} \\
\midrule
Implementation & Which weights, processor, score map and text define the classifier? & Prompt-family effects occur on identical images; a model name alone is incomplete. \\
\midrule
Population and endpoint & Which setting and reference labels support the result? & Cohort ranking changes, and dataset TB labels do not share a verified common microbiological standard. \\
\midrule
Negative spectrum & Are controls healthy, broadly abnormal or a named competing disease? & AUROC falls for all models on sick controls; named alternatives expose larger medical-model failures. \\
\midrule
Score interpretation & Which prevalence and reliability measure support the claim? & Class weighting changes Brier comparisons without changing score ordering. \\
\midrule
Operating rule & Does an unchanged source threshold retain sensitivity and acceptable specificity? & Only four of sixteen sensitivity constraints survive by point estimate; retention can coexist with near-zero specificity. \\
\midrule
Uncertainty and provenance & What is resampled, what remains fixed, and what exposure is known? & Image-level intervals, seed variability and documented pretraining exposure support different inferences. \\
\bottomrule
\end{tabular}
\end{table}

This reporting emphasis aligns with established guidance on intended use, reference standards, data provenance and assessment of deployment risks \citep{tejani2024claim,sounderajah2025stardai,lekadir2025futureai}. It also supports recurring local validation rather than certification from one external score \citep{youssef2023external}. A future clinical study would need a representative target population, a justified reference standard, a prespecified action pathway and assessment of downstream benefit and harm. None of those requirements follows automatically from high retrospective discrimination.

\subsection{Limitations and priorities for further work}

This is a single-task demonstration of a more general evaluation risk. All experiments concern chest X-ray TB labels, so the results do not establish the same effects for other diseases, modalities or clinical actions. Montgomery is small, and the named VinDr controls also have limited sample sizes. The study does not measure demographic fairness or reader--model interaction. Differences between cohorts combine acquisition, spectrum and label variation; their individual causal contributions remain unidentified.

The reference labels also limit clinical interpretation. VinDr-CXR uses radiologist annotations, and we do not harmonize all datasets to a common microbiological TB endpoint. Diagnoses can co-occur. The below-0.5 named-disease estimates characterize relative ranking under the fixed classifier, not an observed rate of clinical misdiagnosis. A prospective screening study could yield a different disease spectrum and different error costs.

Pretraining independence is incompletely observable. CheXficient has documented VinDr-CXR exposure, while other models include training sources whose full membership cannot be reconstructed here. The cross-dataset hash screen omits VinDr-CXR, lacks adjudication and does not test every kind of near-duplicate or patient-level relationship. Byte-level non-matches do not exclude re-encoded copies. The exclusion sensitivity is intentionally conservative and cannot certify a leakage-free benchmark.

Uncertainty has several layers. Image resampling does not capture unknown patient clustering, site sampling or repeated clinical deployments. The original and later blocks use different bootstrap designs, as documented in Section~\ref{sec:statistics}. VinDr threshold intervals hold the source threshold fixed and therefore omit its estimation uncertainty. Five training seeds quantify only a limited component of supervised variability on one split. The positive conditional ensemble interval after candidate exclusion is not evidence that retraining will consistently improve an independently trained model.

Implementation choices further bound reproducibility. The common relative score for MedSigLIP differs from an independently interpreted sigmoid output, and its resize implementation differs from the developer's reference evaluation. Some original checkpoint records lack immutable repository revisions. These qualifications reinforce the need to attach results to the implemented system rather than to an abstract checkpoint identity.

The next scientific priorities are a truly independent clinical cohort, harmonized TB reference standards and prespecified clinically plausible negative spectra. Patient-linked validation would support more appropriate uncertainty estimates. Target recalibration or adaptation should then be evaluated as explicit additional interventions, with separate local development and validation data. These studies could determine whether the identified dependencies remain consequential under actual screening conditions.

\section{Conclusions}
\label{sec:conclusion}

Across four vision-language models and four chest X-ray datasets, favorable discrimination does not consistently preserve model ranking, score reliability or a source-derived screening constraint. Paired prompt and negative-spectrum analyses reveal changes that cross-dataset AUROC alone cannot explain. A near-perfect supervised source result and conservative overlap-candidate exclusion do not remove the external gap. The evidence is specific to this TB-screening audit, but it illustrates a general risk in attributing clinical portability to a checkpoint without its evaluation conditions. A defensible claim identifies the classifier, population, negative spectrum, prevalence assumption, operating rule and uncertainty that support it.

\section*{Data and Model Sources}

The study uses the Montgomery and Shenzhen collections \citep{jaeger2014tb}, the labeled TBX11K training and validation splits \citep{liu2020tbx11k}, and the VinDr-CXR version 1.0.0 test set and annotations \citep{nguyen2022vindr,nguyen2021vindrphysionet}. VinDr-CXR access is governed by PhysioNet credentialing and its data-use conditions \citep{goldberger2000physionet}. Model sources and fixed scoring choices appear in Sections~\ref{sec:implementation} and~\ref{sec:scoring}.

\FloatBarrier
\bibliographystyle{unsrtnat}
\bibliography{references}
\end{document}